\documentclass[journal]{IEEEtran}
\usepackage{amsmath,amsfonts}
\usepackage{algorithmic}
\usepackage{algorithm}
\usepackage{array}
\usepackage[caption=false,font=normalsize,labelfont=sf,textfont=sf]{subfig}
\usepackage{textcomp}
\usepackage{stfloats}
\usepackage{url}
\usepackage{verbatim}
\usepackage{graphicx}
\usepackage{cite}
\usepackage{xcolor}
\usepackage{threeparttable}
\usepackage{makecell}
\usepackage{tablefootnote}
\usepackage{wasysym}
\usepackage{arydshln}
\usepackage{longtable}
\usepackage{multirow}
\usepackage{hyperref}
\usepackage{hhline}
\usepackage{amssymb}
\usepackage{pifont}
\usepackage{utfsym}
\usepackage{adjustbox}
\usepackage{comment}
\newcommand{\xmark}{\scalebox{0.75}{\usym{2613}}}%

\newcommand{\smallsim}{\smallsym{\mathrel}{\sim}}

\makeatletter
\newcommand{\smallsym}[2]{#1{\mathpalette\make@small@sym{#2}}}
\newcommand{\make@small@sym}[2]{%
  \vcenter{\hbox{$\m@th\downgrade@style#1#2$}}%
}
\newcommand{\downgrade@style}[1]{%
  \ifx#1\displaystyle\scriptstyle\else
    \ifx#1\textstyle\scriptstyle\else
      \scriptscriptstyle
  \fi\fi
}
\makeatother

\begin{document}

\title{Autonomous Driving Small-Scale Cars: A Survey of Recent Development}

\author{Dianzhao Li, Paul Auerbach, and Ostap Okhrin

\thanks{Dianzhao Li and Ostap Okhrin are with the Chair of Econometrics and Statistics, esp. in the Transport Sector, Technische Universität Dresden, Dresden, 01187, Germany and Center for Scalable Data Analytics and Artificial Intelligence (ScaDS.AI) Dresden/Leipzig, Germany. {(E-mails: dianzhao.li@tu-dresden.de; ostap.okhrin@tu-dresden.de)}\\
Paul Auerbach is with Barkhausen Institut gGmbH, 01067 Dresden, Germany. {(E-mail: paul.auerbach@barkhauseninstitut.org)}
}
}



\maketitle

\begin{abstract}
While engaging with the unfolding revolution in autonomous driving, a challenge presents itself, how can we effectively raise awareness within society about this transformative trend? While full-scale autonomous driving vehicles often come with a hefty price tag, the emergence of scaled-down, small-scale car platforms offers a compelling alternative. These miniature vehicles are designed to perform predefined tasks and challenges, equipped with onboard sensors, processing units, and control actuators. These platforms not only serve as valuable educational tools for the broader public and young generations but also function as robust research platforms, contributing significantly to the ongoing advancements in autonomous driving technology. This survey outlines various small-scale car platforms, categorizing them and detailing the research advancements accomplished through their usage. The conclusion provides proposals for promising future directions in the field. 
\end{abstract}


\section{Introduction}

\IEEEPARstart{O}{ver} the past few decades, extensive research on autonomous driving (AD) has been conducted by both academic communities and industry stakeholders. However, the fruition of fully functional Level 5 autonomous driving systems, which entails the availability of completely autonomous vehicles in the mass market, is anticipated to materialize in the coming decades \cite{khan2022level, chen2022milestones}. The question arises: \emph{Can we proactively prepare our society for the oncoming fully autonomous driving?} Despite assertions from researchers that autonomous vehicles (AVs) will mitigate human error and enhance safety compared to human drivers, public apprehensions persist regarding the ethical and safety dimensions of AVs \cite{fraedrich2016societal, rezaei2020examining}. Encouragingly, individuals with prior experience with AVs and younger generations exhibit a more optimistic stance toward these technologies \cite{othman2021public}. Hence, the affirmative response to this question is clear. Given that those who will further develop the AD systems are currently in high school, and those who will coexist with AVs are presently in elementary schools, the optimal preparation involves providing opportunities for the public to engage with AD systems now. To this end, a small-scale car platform emerges as an ideal choice, serving educational purposes and facilitating researchers in testing their autonomous systems on a tangible platform. \par

\begin{figure*}[!t]
    \centering
    \includegraphics[width=0.8\textwidth]{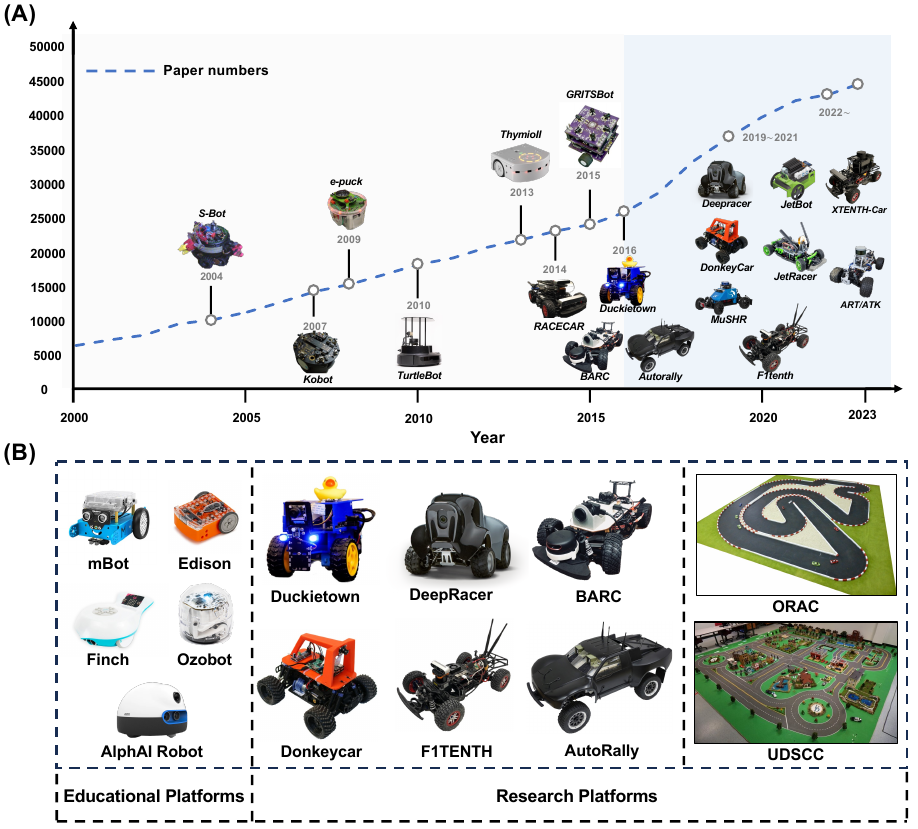}
    \caption{Illustration of the development and current states of small-scale car platforms, each depicted platform image sourced from its respective paper or website. (A) Based on published studies each year on Google Scholar with the search terms: "small-scale car" or "robot car", the research on small-scale cars has seen substantial growth in the number of papers over the years. In the early 2000s, projects like s-bot \cite{mondada2004swarm}, e-puck \cite{mondada2009puck}, and TurtleBot emerged. Starting around 2016, with the introduction of Duckietown \cite{paull2017duckietown}, BARC \cite{gonzales2016autonomous}, and Autorally \cite{goldfain2019autorally}, there was a significant increase in research papers. This trend continued with the development of projects like DeepRacer \cite{balaji2020deepracer}, Donkeycar, and F1TENTH \cite{o2020f1tenth}. More computationally advanced small-scale cars have been introduced in recent years, such as ART/ATK \cite{elmquist2022software} and XTENTH-CAR \cite{sivashangaran2023xtenth}. (B) Examples of small-scale car platforms, categorized into educational platforms and research platforms, including multiple vehicle setups such as ORAC \cite{liniger2015optimization} and UDSSC \cite{stager2018scaled}.}
    \label{fig:Robot car platforms}
\end{figure*}

Undoubtedly, research in AD research for full-scale cars holds promise. Despite the supportive environments for developing and testing AVs in various countries and regions, strict regulations persist due to concerns about safety, security, and public trust \cite{fagnant2015preparing, pattinson2020legal,taeihagh2019governing}. These regulations often limit the ability to drive or test AVs on public roads, which consequently restricts the involvement of smaller research institutions or individual researchers in AD system development. As a solution, the use of small-scale car platforms offers a more cost-effective and accessible alternative for the public and research communities to engage with AD technologies. More specifically, small-scale car platforms can serve two main purposes: educational tools for students and research tools for AD researchers. In educational settings, small-scale cars prove to be an excellent choice for schools, especially with science, technology, engineering, and mathematics (STEM) focus, offering students their initial exposure and hands-on experience with AD while simultaneously fostering increased public awareness. On the research front, the availability of such research platforms lowers entry barriers, inviting a broader spectrum of researchers into AD exploration. As evidenced in Fig. \ref{fig:Robot car platforms}(A), based on published studies each year on Google Scholar with the search terms: "small-scale car" or "robot car" from 2000 onward, the quantity of research papers focused on small-scale car platforms has experienced a substantial surge in the past two decades. Following a relatively gradual incline before 2016, this field has undergone a notable expansion, particularly catalyzed by the introduction of a series of well-known platforms. It is worth mentioning that AD research for small-scale cars is the ultimate goal of these small-scale car platforms, not the transformation of the techniques into real-size vehicles. Various major small-scale car platforms claim that these platforms are designed to be accessible and inexpensive, aimed at fostering educational and research activities \cite{paull2017duckietown, balaji2020deepracer, goldfain2019autorally,o2020f1tenth, gonzales2016autonomous}. For example, as stated by Duckietown platform \cite{paull2017duckietown}, their mission is \emph{Learning robotics and AI made tangible, accessible, and fun!}  They want to provide a tangible, accessible, and inclusive tool for a broader range of society to engage with robots and AV. Similarly, another major platform DeepRacer \cite{balaji2020deepracer}, provided by Amazon Web Services, also tries to offer developers of all skill levels for hands-on experience with fully autonomous 1/18th scale race cars with machine learning (ML). Another platform, F1TENTH's mission is to \emph{foster interest, excitement, and critical thinking about the increasingly ubiquitous field of autonomous systems} \cite{o2020f1tenth}. In comparison to full-scale AVs, small-scale car platforms employ more lightweight computation units and sensors for economic reasons. These platforms primarily use mature AV techniques from full-scale cars, albeit under simplified conditions due to current hardware limitations. Nonetheless, they play a crucial role in engaging undergraduate students and early career researchers in AV technologies, inspiring them, and contributing to the growth of the AV research community. This, in turn, indirectly advances real-world AV research. \par

\subsection{Contributions}

Despite the importance and widespread use of small-scale car platforms, no research currently provides a comprehensive overview of these platforms, including their hardware configurations, software frameworks, and benchmarking of driving tasks. In \cite{caleffi2023systematic}, the existing literature on small-scale cars is summarized; however, it lacks comprehensiveness and does not include a comparative analysis. More recently, \cite{mokhtarian2024survey} discussed existing platforms, but their focus was limited to the platforms themselves, without addressing the techniques used to advance small-scale car AD research. Identifying this gap, we aim to advance ongoing research initiatives and enhance public awareness by conducting a comprehensive survey of existing small-scale platforms for educational and research purposes. Additionally, we seek to benchmark the AD tasks accomplished by these platforms. To the best of our knowledge, this survey represents the first of its kind to provide such an extensive analysis.  The contributions of this work are as follows:

\begin{itemize}
    \item \textbf{Comprehensive Review of Platforms}: We thoroughly review the most widely available small-scale car platforms, comparing their hardware and software configurations to provide readers with an in-depth understanding and a guide for selecting or building their own platforms.
    \item \textbf{Benchmarking Autonomous Driving Tasks}: By analyzing over 250 research papers, we categorize the driving tasks for small-scale cars into two primary pipelines—modular pipelines and end-to-end pipelines. We discuss in detail the techniques employed in each module within these pipelines.
    \item \textbf{Proposed Future Directions}: After examining the existing techniques, we propose potential improvements for both the software and hardware aspects. On the software side, we suggest techniques to enhance driving behaviors, while on the hardware side, we recommend platform improvements to facilitate more advanced research.
\end{itemize}

The rest of this paper is organized as follows: In Section \ref{sec:platforms}, we examine commonly employed small-scale car platforms, accompanied by insights into their simulators. Following that in Section \ref{sec:sensors}, we inspect the sensor setups employed in these platforms. Section \ref{sec:Task} provides a comprehensive overview of the AD tasks accomplished within the research community. Section \ref{sec:Future} outlines promising directions for future developments, and we draw our conclusions in Section \ref{sec:Conclusion}.

\section{Platforms}
\label{sec:platforms}

For full-scale cars, the standardized dynamics and sizes are mandated by road regulations. In contrast, smaller-scale cars, designed without regulatory constraints, are crafted for diverse research purposes by various groups of researchers and enthusiasts. Recognizing these disparities, this survey establishes a clear definition for AD robot cars, setting them apart from other autonomous robots. In this context, an AD small-scale car platform refers to a miniature vehicle equipped with technologies enabling autonomous operation. This includes sensors, processing units, control actuators, and often a set of predefined tasks or challenges for testing and development. In terms of size, namely the scale factors, there is considerable variability among different platforms. Platforms intended to carry a lot of sensors and for outdoor operation can reach almost 1 meter in length \cite{goldfain2019autorally}. On the other end of the spectrum, platforms developed for investigating swarm dynamics are notably compact, with some as small as 3.3x3.3cm \cite{rubenstein2012kilobot}. The most common form factor is the 1/10th scale, representing a model approximately one-tenth the size of a full-scale car. This scale is frequently employed in hobby Remote-Control (RC) vehicles, forming the foundation for numerous platforms \cite{donkeycar, o2020f1tenth, karaman2017project, gonzales2016autonomous, srinivasa2019mushr, vincke2021open, berlin2020autominy, pohlmann2022ros2, sivashangaran2023xtenth}.
Regarding the dynamic system, full-scale cars commonly use Ackermann steering mechanics, rotating the front axle to facilitate left or right turns \cite{gillespie2021fundamentals}. However, due to the mechanical complexity of this system and its limitations in certain situations, small-scale car platforms initially favored the use of a differential steering system. In this system, the left and right sets of wheels are driven independently, enabling turns by driving one set of wheels faster than the other. Nevertheless, as illustrated in Fig. \ref{fig:Robot car platforms}, recent advancements in this field have seen the increased adoption of the more realistic Ackermann steering system in small-scale car platforms.

\subsection{Hardware Platforms}
\label{subsec:platform}

Before we introduce the different platforms, we categorize small-scale car platforms into distinct groups based on their target users and complexity.

\subsubsection{Educational Platforms}
\label{subsubsec:educational_platforms}
First are the educational platforms predominantly accessible in the commercial market, as shown in Table \ref{tab:platforms_educations}, including Makeblock mBot, Edison, AlphAI Robot,
TinkerGen MARK, and Robolink Zumi. These platforms are carefully crafted to engage students ranging from early childhood to elementary school and even graduate programs in the interactive exploration of autonomous cars. They offer users comprehensive tools, enabling them to construct effortlessly and program robot cars capable of executing diverse tasks. These platforms are equipped with basic sensors such as line sensors, distance sensors, and occasionally cameras. With limited processing power, most rely on microcontrollers, they are suitable only for simple tasks like lane keeping or car following. However, their support for simple frameworks and visual programming languages makes them ideal for undergraduate education. Many also come with pre-made courses designed to teach students the fundamentals of their operation. Their low cost and commercial availability further enhance their accessibility for educational purposes. Despite these advantages, these platforms are not well-suited for researchers exploring autonomous or connected driving. Their closed-source hardware and software limit extensibility, and most lack networking capabilities for external control or integration into multi-vehicle scenarios. Additionally, their differential-drive steering design does not accurately replicate real car dynamics, making them inadequate for serious investigations into realistic vehicular behavior. \par 

\begin{table}
\tiny
\centering
    \resizebox{\linewidth}{!}{%
    \begin{threeparttable}
    \caption{An overview of small-scale car platforms for educational purposes.}
    \begin{tabular}{|l|c|l|l|c|c|}
    \hline
    \textbf{Platforms} & \textbf{Size}  & \textbf{Sensors} & \textbf{Programming} & \textbf{Runtime} & \textbf{\makecell[c]{Price (USD)}} \\ \hline
        \textbf{Thymio} \cite{riedo2013thymio} & 110x110mm  & \makecell[l]{IR sensor, Accelerometer, \\Microphone, Thermistor} & VPL &2h& 270\\ \hline
        \textbf{\makecell[l]{Makeblock \\mBot}} & -- & \makecell[l]{Ultrasonic Sensor, IR sensor, \\ Line tracking sensor}  & mBlock & 1h &190 \\ \hline
        \textbf{Edison} & 80x80x40mm & IR sensor, Line tracking sensor  & \makecell[l]{EdBlock \\ EdPy} & -- &60\\ \hline
        \textbf{\makecell[l]{AlphAI\\Robot}} & -- & \makecell[l]{Camera, Ultrasonic Sensor, \\ Line tracking sensor}  & Python & -- & 270 \\ \hline   
        \textbf{\makecell[l]{Ozobot\\Evo}} & 32mm \diameter &  IR sensor, Speaker & OzoBlockly& 1h & 160 \\ \hline     
        \textbf{Finch Robot} & -- & \makecell[l]{Encoder, Ultrasonic sensor, \\Line tracking sensor, IR sensor, \\Speaker} & MicroBit& 7h & 170\\ \hline 
        \textbf{\makecell[l]{TinkerGen\\MARK}} & 200x185x92mm & Camera, Ultrasonic Sensor & MicroPython & -- & 220\\ \hline
        \textbf{\makecell[l]{Robolink\\Zumi}} & 95x67x70mm & Camera, IR sensor & Python& -- & 190\\ \hline
    \end{tabular}
    \label{tab:platforms_educations}
    \end{threeparttable}}
\end{table}

\subsubsection{Commercial Research Platforms}

The subsequent tier has a more sophisticated system, affording users enhanced opportunities to innovate and develop new functionalities within the platform. These platforms also mostly offer support for more advanced frameworks like Robot Operating System (ROS) and standard programming languages such as Python to incorporate them in bigger research projects. Their inclusion of networking capabilities also aids in incorporating them in diverse environments. The typical platforms include e-puck \cite{mondada2009puck}, Pheeno \cite{wilson2016pheeno}, Thymio \cite{riedo2013thymio}, MarXbot \cite{bonani2010marxbot}, Turtlebot, and Duckietown \cite{paull2017duckietown}. The e-puck, Pheeno, Thymio, and MarXbot are frequently used in swarm robotics research due to their compact size and reliance on basic sensors. The Duckietown ecosystem originated in 2016 at MIT as an educational tool for instructing students in the realms of autonomous and connected driving but developed into a research platform for all levels of researchers. Comprising cost-effective vehicles known as Duckiebots, each unit is outfitted with either a Raspberry Pi or a Jetson Nano as its computing unit. Sensor-wise, the Duckiebots employ an RGB Raspberry Pi camera, an inertial measurement unit (IMU), wheel encoders, and a front-facing distance sensor. The platform also facilitates the incorporation of a central localization system through "watchtowers." Propelled by two DC motors on each of the two driven wheels, the cars adhere to a differential drive pattern. Duckietown is commercially available and widely used for tasks such as lane keeping using the RGB camera \cite{rosman2017hybrid, perez2020interactive}, as well as obstacle and traffic sign detection \cite{nubert2018developing, purwanto2021autonomous}. Its support for diverse frameworks and sensors, along with its open-source nature, makes it applicable to various target groups and tasks. Comprehensive documentation and a large community further enhance its adaptability and integration into different environments. The differential-drive steering setup in Duckietown, similar to educational platforms, is a significant limitation. Another commercially available platform is the AWS DeepRacer \cite{balaji2020deepracer} by Amazon Web Services. This platform also relies on an RGB camera for sensory input, supplemented by an upgrade kit featuring a second camera and a LiDAR sensor. The execution of learned strategies is managed by a small compute board equipped with an Intel Atom CPU. With Ackermann steering implementation, DeepRacer is also primarily used for tasks like lane keeping \cite{sandha2021sim2real, chiba2021effectiveness} and obstacle avoidance \cite{mccalip2023reinforcement, mitchell2020multi}. While the platform offers open-source software to interface with its hardware, its heavy reliance on Amazon Web Services limits extensibility beyond the intended use of the manufacturer. However, AWS provides extensive documentation, including detailed courses on setup and operation within its software ecosystem. \par

\subsubsection{Open Source Platforms}
\label{subsubsec:opensource_platforms}

For users seeking open-source platforms and aiming to build a system from the ground up, two popular options are Donkeycar\cite{donkeycar} and F1TENTH \cite{o2020f1tenth}. Donkeycar is a community-driven platform that provides detailed instructions for constructing a customizable vehicle. Its open-source nature allows significant flexibility in sensor configurations and system design. The base platform is built on a 1/10th-scale RC car with Ackermann steering, equipped with a Raspberry Pi or Jetson Nano as the compute unit and a Raspberry Pi camera as the primary sensor.  Users can enhance the platform by adding sensors such as IMUs, encoders, and even LiDAR. Given its open-source framework, users can extend the platform's functionality according to their needs. Originally designed for autonomous racing, Donkeycar is commonly used for tasks like lane keeping \cite{viitala2021learning, zhou2021deep} and obstacle avoidance \cite{yun2021virtualization}. However, its versatility extends to a wide range of applications due to its open-source framework. The community behind the project also offers comprehensive documentation on building and using the platform. Its Python-based custom framework supports integration with other systems, such as ROS, further expanding its functionality. While primarily designed for single-vehicle use, the platform could be extended to enable coordination among multiple vehicles. Despite its advantages, the complexity and the need for manual assembly and configuration may deter researchers seeking out-of-the-box solutions. F1TENTH, like Donkeycar, is based on a 1/10th-scale RC car with Ackermann steering. It features a Nvidia Jetson as its compute unit and a 2D LiDAR as the primary sensor. Additional hardware includes an IMU, odometry data provided by the VESC motor controller, and an optional RGB camera. With its high-speed capable base platform, F1TENTH is primarily utilized for autonomous racing tasks \cite{sinha2020formulazero, tuatulea2020design}, though it is also applied in fundamental tasks like lane keeping \cite{ivanov2020case} and obstacle avoidance \cite{bulsara2020obstacle, verma2021implementation}. Thanks to its relatively powerful compute unit, F1TENTH can support advanced applications, including ML-based decision-making. The big advantage of these two platforms lies in their open-source nature and comprehensive online documentation, enabling extensive customization, such as incorporating additional sensors or connecting to simulators. However, this flexibility comes with the challenge of increased complexity, as users must build and configure the platforms from scratch, an approach that may not be suitable for all researchers.\par

\subsubsection{Outdoor Platforms}

For outdoor applications, the Autorally \cite{goldfain2019autorally} platform takes center stage, built on an off-the-shelf RC car platform. Unlike smaller platforms such as Donkeycar and F1TENTH, AutoRally uses a larger 1/5th-scale model car, allowing it to accommodate more powerful hardware. Its compute unit is a consumer-grade computer mainboard equipped with an Intel i7 CPU, enabling advanced processing capabilities. The platform features a sophisticated sensor suite, including two industrial synchronized RGB cameras for stereo imaging, an IMU, magnetic encoders for odometry, and a GPS receiver. Due to its substantial size and high-performance components, Autorally is mainly employed for racing tasks \cite{williams2017information, williams2016aggressive, you2019high}. These tasks often utilize various sensor fusion strategies, such as combining GPS and IMU data \cite{wagener2019online, lee2020perceptual} or integrating camera and IMU data \cite{drews2019vision, drews2017aggressive}. While the diverse set of sensors and substantial computing power allow the platform to be used for different tasks, its large size and high-speed capabilities necessitate an outdoor environment, making it unsuitable for smaller lab setups. The project provides instructions for constructing the vehicle, setting up the software, and integrating it into the ROS framework. However, the lack of comprehensive documentation may present challenges for users attempting to fully leverage its potential. \par

\subsubsection{Multi Vehicle Platforms}

The significance of a small-scale smart city platform, accommodating multiple vehicles, is underscored as it plays a crucial role in achieving harmony and efficient collaboration of a fully autonomous driving world. To this end, the University of Delaware Smart Scaled City (UDSSC) \cite{stager2018scaled}, developed for education and research on connected and autonomous driving, emerges as a distinctive contribution. UDSSC features a scaled-down urban environment with diverse traffic scenarios, including intersections and roundabouts, and uses Micro Connected and Automated Vehicles (MCAVs). These vehicles are built on the Pololu Zumo platform and feature a differential-drive DC motor setup with encoders, an IMU, and an Arduino microcontroller. Each MCAV is further equipped with a Raspberry Pi as its compute unit, a line-following sensor, and a front-facing distance sensor. Notably, the absence of onboard cameras or LiDAR necessitates the use of a centralized localization system for precise positioning. The platform is tailored for cooperative and connected driving tasks, such as merging \cite{stager2018scaled} and roundabout navigation \cite{chalaki2020experimental, zayas2022digital, jang2019simulation}. However, it is constrained by several limitations. The reliance on an external localization system ties the platform to a specific physical testbed, as relocating the system requires substantial effort. Additionally, the lack of onboard sensors for localization restricts its use to studies on multi-vehicle interactions. Unfortunately, the UDSSC platform is not open source; its hardware and software are proprietary, and it is not available for purchase. Moreover, the absence of detailed documentation limits its usability. As a result, the platform primarily serves as inspiration for developing custom solutions with similar capabilities.\par

For the sake of brevity, the details of commonly used platforms have been summarized in tables. Table \ref{tab:platforms_hw} presents a comparative hardware analysis of various platforms, highlighting key characteristics such as steering dynamics, actuators, installed sensors, and prices, whether for crafting or purchasing a platform. Table \ref{tab:platforms_sw} focuses on the software aspect, detailing the software frameworks employed and, where applicable, the simulation tools integrated into each platform. Additionally, the primary focus tasks of each platform are provided as a reference for those interested in exploring specific driving behaviors.

\begin{table*}
\tiny
\centering
    \begin{threeparttable}
    \caption{Small-scale car platforms: an overview of the hardware setup.}
    \begin{tabular}{|l|l|l|l|l|l|c|c|c|}
    \hline
    \textbf{Platforms} & \textbf{Size} & \textbf{\makecell[l]{Vehicle \\ Dynamics}} & \textbf{Actuator} & \textbf{Sensors} & \textbf{Computation Unit} & \textbf{Runtime} & \textbf{\makecell[c]{Commercial\\available}} & \textbf{\makecell[c]{Price (USD)}} \\ \hline
        \textbf{AutoRally} \cite{goldfain2019autorally}& 1/5th & Ackermann & Two servo motors & \makecell[l]{Camera, IMU, GPS, \\Hall-effect sensor} & \makecell[l]{Intel i7-6700 \\ Nvidia GTX-750ti SC} & \textless1h & \xmark\footnotemark[7] & 10k\\ \hline
        \textbf{ART/ATK} \cite{elmquist2022software} & 1/6th & Ackermann  & \makecell[l]{One brushless DC motor\\ One servo motor} & \makecell[l]{Camera, 3D LiDAR} & Jetson Xavier NX  & --  & \xmark 
 & --\\ \hline
        \textbf{BARC} \cite{gonzales2016autonomous} & 1/10th & Ackermann & \makecell[l]{One brushless DC motor} & \makecell[l]{Camera, LiDAR, IMU, GPS}  & ODROID-XU4 & -- & \xmark & -- \\ \hline
        \textbf{Donkeycar} \footnotemark[1] & \makecell[l]{1/10th\\1/16th } & Ackermann & One brushed/brushless DC motor & \makecell[l]{Camera, LiDAR, IMU, Encoder} & RPi/Jetson Nano  & $\checked$ & \xmark\footnotemark[7] & 350\\ \hline
        \textbf{F1TENTH} \cite{o2020f1tenth} & 1/10th & Ackermann & One brushless DC motor& \makecell[l]{Camera, LiDAR, IMU} & Jetson TX2 & \textless1h& \xmark\footnotemark[7] & 3800\\  \hline
        \textbf{RACECAR(MIT)} \cite{karaman2017project} & 1/10th & Ackermann & \makecell[l]{One brushless DC motor\\ One servo motor} & \makecell[l]{Camera, LiDAR, IMU, Encoder} & Jetson Tegra X1 & -- & \xmark\footnotemark[7] & 2600\\ \hline
        \textbf{MuSHR} \cite{srinivasa2019mushr} & 1/10th & Ackermann &\makecell[l]{One brushless DC motor\\ One servo motor}& \makecell[l]{Camera, LiDAR, IMU,\\ Bump sensor} & Jetson Nano & -- & \xmark\footnotemark[7] & 900\\ \hline
        \textbf{Qcar} \footnotemark[2]  & 1/10th & Ackermann & \makecell[l]{One brushless DC motor}& \makecell[l]{Camera, LiDAR, IMU, Encoder,\\Microphone} & Jetson TX2 & 30m$\smallsim$2h & $\checked$ & --\\ \hline
        \textbf{Autominy} \cite{berlin2020autominy} & 1/10th & Ackermann & One brushless DC Servomotor & \makecell[l]{Camera, LiDAR, IMU, Encoder} & Intel NUC & -- & \xmark & --\\ \hline
        \textbf{JetRacer} \footnotemark[3]& \makecell[l]{1/10th\\1/18th} & Ackermann & \makecell[l]{One brushed DC motor} & Camera & Jetson Nano & --  & $\checked$ & 600\\ \hline
        \textbf{Autonomouscar} \cite{vincke2021open}& 1/10th & Ackermann & One brushed DC motor & \makecell[l]{Camera, LiDAR, IMU, Encoder, \\ ToF Sensor, Indoor GPS}  & RPi 4 & --& \xmark & --\\ \hline
        \textbf{CoRoLa} \cite{pohlmann2022ros2} & 1/10th & Ackermann &\makecell[l]{One brushless DC motor\\ One servo motor} & \makecell[l]{Camera, Encoder, \\Ultrasonic sensor}& RPi 4 & -- & \xmark & --\\ \hline
        \textbf{AutoDRIVE} \cite{samak2023autodrive} & 1/14th & Ackermann & \makecell[l]{Two DC geared motors} & \makecell[l]{Camera, LiDAR, IMU, Encoder, \\Indoor GPS} & Jetson Nano & -- & \xmark & --\\ \hline
        \textbf{PiRacer} \footnotemark[4] & 1/16th & Ackermann &\makecell[l]{Two DC brushed motors} & Camera & RPi 4 & -- & $\checked$ & 250\\ \hline
        \textbf{Duckietown} \cite{paull2017duckietown} & 34x15x23cm & Differential & \makecell[l]{Two DC geared motors} & \makecell[l]{Camera, IMU, Ultrasonic sensor} & RPi 2/Jetson Nano & 2$\smallsim$6h & $\checked$ & 450\\ \hline
        \textbf{DeepRacer} \cite{balaji2020deepracer} & 1/18th & Ackermann &\makecell[l]{One brushless DC motor\\ One servo motor}& Camera, LiDAR, IMU & Intel Atom & $\smallsim$6h & $\checked$ & 400\\ \hline
        \textbf{µcar} \cite{kloock2021cyber} & 1/18th & Ackermann &\makecell[l]{One brushless DC motor\\ One servo motor}& IMU, Encoder & RPi Zero W &$\smallsim$6h & \xmark & --\\ \hline
        \textbf{UDSSC MCAV} \cite{stager2018scaled} & 1/25th & Ackermann & One geared DC motor & \makecell[l]{IMU, line following, IR sensor} & RPi 3 &90m& \xmark & --\\ \hline
        \textbf{Chronos} \cite{carron2023chronos} & 1/28th & Ackermann & \makecell[l]{Brush motor with gearbox}& IMU, Encoder & Espressif ESP32 &30m$\smallsim$1h & \xmark & --\\ \hline
        \textbf{Go-CHART} \cite{kannapiran2020go} & 1/28th & Differential & \makecell[l]{Four micro metal gear motors} & \makecell[l]{Camera, LiDAR, Bump sensor} & RPi 3 & 	$\smallsim$1h& \xmark & --\\ \hline  
        \textbf{Cambridge Minicar} \cite{hyldmar2019fleet} & 75x81x197mm & Ackermann & -- & Indoor positioning system\footnotemark[8] & RPi Zero & 2h & \xmark & --\\ \hline
        \textbf{ORCA Racer} \cite{liniger2015optimization} & 1/43th & Ackermann & -- & IMU, Indoor positioning system\footnotemark[8] & ARM Cortex M4 & 20m & \xmark & --\\ \hline
        \textbf{Epuck} \cite{mondada2009puck} &  70mm \diameter & Differential & Two stepper motors& \makecell[l]{Camera, IMU, IR sensor, Speaker, \\Microphone}& STM32F407 & $\smallsim$3h & $\checked$ & 1000 \\ \hline
        \textbf{Turtlebot3} \footnotemark[5]  &14x18x19cm & Differential & Two servomotors& Camera, LiDAR, IR sensor & RPi 4 & $\smallsim$2.5h & $\checked$ & 1200\\ \hline
        \textbf{Kilobot} \cite{rubenstein2012kilobot} & 33mm \diameter & Vibration & Two vibration motors & IR sensor & Atmega 328 & 3$\smallsim$10h& $\checked$ & 15\\ \hline
        \textbf{GRITSBot} \cite{pickem2015gritsbot} & 31x30mm & Differential & Two stepper motors& \makecell[l]{IMU, IR sensor} & Atmega 328  & 1$\smallsim$5h & \xmark  & --\\ \hline
        \textbf{HydraOne} \cite{wang2019hydraone} & 27x32cm & Omni &Two encoder motors& \makecell[l]{Camera, 3D LiDAR, Encoders} & Jetson TX2  & -- & \xmark\footnotemark[7] & 7200\\ 
        \hline
        \textbf{Pheeno} \cite{wilson2016pheeno} &  10cm \diameter & Differential & Two micro gear motors& \makecell[l]{Camera, IMU, Encoder, IR sensor}& \makecell[l]{ATmega328P \\ ARM Cortex-A7} & 5h & \xmark\footnotemark[7] & 270 \\ \hline
        \textbf{Thymio} \cite{riedo2013thymio} & 11x11cm & Differential & Two DC motors& \makecell[l]{IR sensor, Accelerometer, \\Microphone, Thermistor} & PIC24F &2h & $\checked$ & 270\\ \hline
        \textbf{MarXbot} \cite{bonani2010marxbot} & 17cm \diameter & Differential & -- & \makecell[l]{Camera, IMU, IR sensor, \\ 2D force sensor}  & ARM 11 processor & -- & \xmark & --\\ \hline
        \textbf{WolfBot} \cite{betthauser2014wolfbot} & 17.5cm \diameter & Omni & --& \makecell[l]{Camera, IMU, IR sensor, \\ Microphone} & BeagleBone  & --  & \xmark & -- \\ \hline
        \textbf{LabRAT} \cite{robinette2009labrat} & - & Differential &  Two
DC gearmotors & \makecell[l]{IR sensor} & Atmega324p & 3h & \xmark & -- \\ \hline
        \textbf{Jetbot} \footnotemark[6] & - & Differential & Two TT motors &  Camera, IMU & Jetson Nano & 2$\smallsim$3h& $\checked$ & 250\\ \hline
    \end{tabular}
    \begin{tablenotes}[para,flushleft]
        \tiny
        \item[1.] http://donkeycar.com  
        \item[2.] https://www.quanser.com/products/qcar  
        \item[3.] https://github.com/NVIDIA-AI-IOT/jetracer \\
        \item[4.] https://www.waveshare.com/wiki/PiRacer\_AI\_Kit
        \item[5.] https://www.turtlebot.com  
        \item[6.] https://jetbot.org \\
        \item[7.]Build guides with off-the-shelf parts are provided.
        \item[8.]Motion Capture Systems.\\
      \end{tablenotes}
    \label{tab:platforms_hw}
    \end{threeparttable}
\end{table*}

\begin{table}
\tiny
\centering
\begin{adjustbox}{width=0.5\textwidth}
\begin{threeparttable}
\caption{Software setups and target tasks for different small-scale car platforms.}

\begin{tabular}{|l|l|l|l|c|}
\hline
\textbf{Platforms} & \textbf{\makecell[l]{Software \\ Programming}} & \textbf{\makecell[l]{Simulation \\ Platform}} & \textbf{\makecell[l]{Focused \\ Tasks\footnotemark[1]}} & \textbf{ \makecell[l]{User \\ Manual}} \\ \hline
\textbf{ART/ATK} & ROS2 & Chrono & General & \xmark \\ \hline
\textbf{MuSHR} & ROS & -- & General  & $\checked$ \\ \hline
\textbf{Qcar} & \makecell[l]{ROS, Python,\\ MATLAB} & Gazebo & General & $\xmark$ \\ \hline
\textbf{Autominy} & ROS & Gazebo & General & $\checked$ \\ \hline
\textbf{AutoDRIVE} & ROS & \makecell[l]{AutoDRIVE \\ Simulator} & General & \checked \\ \hline
\textbf{WolfBot} & ROS & MATLAB & General & \xmark \\ \hline
\textbf{HydraOne} & ROS & -- & General & \checked\\ \hline
\textbf{Turtlebot3} & ROS & Gazebo & General & $\checked$ \\ \hline

\textbf{RACECAR(MIT)} & ROS & Gazebo & \makecell[l]{SLAM\\Path Following} & $\checked$ \\ \hline
\textbf{Donkeycar} & Python & Gym-Donkeycar\footnotemark[2] & \makecell[l]{Lane Keeping \\Racing} & $\checked$ \\ \hline
\textbf{Duckietown}  & ROS & \makecell[l]{Gazebo\\Gym-Duckietown\footnotemark[3]} & \makecell[l]{Lane Keeping\\Obstacle Avoidance} & $\checked$\\ \hline
\textbf{PiRacer} & ROS & -- & Lane Keeping & $\checked$ \\ \hline
\textbf{Jetbot} & ROS & Gazebo & \makecell[l]{Lane Keeping\\ Obstacle Avoidance} & $\checked$\\ \hline
\textbf{Autonomouscar} & \makecell[l]{ROS \\ LabVIEW} & -- & \makecell[l]{Lane Keeping \\ Obstacle Avoidance} & \xmark \\ \hline

\textbf{AutoRally} & ROS & Gazebo & Racing & $\checked$ \\ \hline
\textbf{BARC} & ROS & -- & \makecell[l]{Racing \\ Drifting} & \xmark \\ \hline
\textbf{F1TENTH} & ROS & \makecell[l]{Gazebo \\ F1TENTH Gym\footnotemark[4]} & Racing &  $\checked$ \\  \hline
\textbf{JetRacer} & -- & -- & Racing & $\checked$ \\ \hline
\textbf{DeepRacer} & ROS & Gazebo & Racing & $\checked$ \\ \hline
\textbf{ORCA Racer} & C++ & -- & Racing & \xmark \\ \hline

\textbf{Epuck} & ROS & \makecell[l]{Enki, Webots\\ V-REP, ARGoS} & Swarm & $\checked$ \\ \hline
\textbf{Kilobot} & -- & V-REP & Swarm & $\checked$ \\ \hline
\textbf{GRITSBot} & -- & -- & Swarm & \xmark  \\ \hline

\textbf{Pheeno} & Python & -- & Swarm & \checked \\ \hline
\textbf{MarXbot} & ROS & Gazebo & Swarm & \xmark \\ \hline

\textbf{LabRAT} & C &  Player/Stage & Swarm & \xmark \\ \hline 

\textbf{CoRoLa} & ROS2 & \makecell[l]{Based on \\ LGSVL} & Cooperative Driving & \xmark \\ \hline
\textbf{µcar} & \makecell[l]{C++ \\ Matlab} & -- & Cooperative Driving & \checked \\ \hline
\textbf{UDSSC MCAV} & ROS & SUMO & Cooperative Driving & \xmark \\ \hline
\textbf{Chronos} & ROS & -- & Cooperative Driving & \checked \\ \hline
\textbf{Go-CHART} & ROS & -- & Cooperative Driving & \xmark \\ \hline  
\textbf{Cambridge Minicar} & Python & -- & Cooperative Driving & \checked \\ \hline       
\end{tabular}
\begin{tablenotes}[para,flushleft]
\tiny
\item[1] Focused tasks are identified based on the main research paper or proposed by us according to the sensor configuration.\\
\item[2] https://github.com/tawnkramer/gym-donkeycar.\\
\item[3] https://github.com/duckietown/gym-duckietown.\\
\item[4] https://github.com/f1tenth/f1tenth\_gym.
\end{tablenotes}

\label{tab:platforms_sw}
\end{threeparttable}
\end{adjustbox}
\end{table}

\subsection{Simulators}

Before employing the autonomous system in real platforms, a simulation is often used first to test the performance. For small-scale cars, numerous simulators exist to facilitate the training, testing, and evaluation of autonomous systems.\par

First, for the requirements of simulating the vehicle dynamics and control systems, CarMaker and CarSim are widely recognized. These platforms provide environments for assessing various aspects of vehicle behavior, including handling, braking, and acceleration, under diverse driving conditions. For autonomous systems, sensor models are as crucial as physics engines. Gazebo \cite{koenig2004design} is a prominent open-source platform in robotics, offering modular systems for integrating diverse sensors and physics models. Gazebo-based simulations have been developed for small-scale platforms like Turtlebot, Duckietown, DeepRacer, F1TENTH, MuSHR, and Autorally, demonstrating its flexibility and adaptability to different research needs. Other robotic simulators such as ARGoS and V-REP are also used as bases for small-scale car platforms \cite{mondada2009puck, rubenstein2012kilobot}. For macro-scale simulations, SUMO \cite{lopez2018microscopic} is a leading platform for analyzing transportation systems, optimizing road networks, and developing traffic management strategies. It has been used for training traffic control algorithms in multi-vehicle systems, as discussed in \cite{jang2019simulation}. In the realm of more detailed autonomous driving setups, various open-source simulators have gained popularity, with TORCS \cite{wymann2000torcs}, CARLA \cite{dosovitskiy2017carla},  AIRSIM \cite{shah2018airsim}, and LGSVL \cite{rong2020lgsvl} being notable examples. These platforms feature high-fidelity physics engines and extensive sensor support, enabling hyper-realistic simulations of traffic environments. They are particularly effective for training larger small-scale vehicles, such as 1/5th scale models, before real-world deployment \cite{mueller2018driving, meng2019neural, bansal2020combining, li2023modified}. In addition to general-purpose simulators, several platforms are tailored to specific small-scale systems. For example, Gym-Duckietown provides a fast and customizable environment for Duckietown, supporting various driving tasks. The DeepRacer cloud simulator by Amazon offers an online platform for customizing algorithms and testing them within the DeepRacer ecosystem. Donkey-Gym, designed for Donkeycar, uses the Unity game engine for enhanced physics and graphics. F1TENTH Gym is optimized for F1TENTH platforms, offering specialized simulation capabilities. AutoDRIVE simulator provides a comprehensive environment for various driving tasks for AutoDRIVE cars. For multi-agent systems, platforms like MADRaS \cite{santara2021madras} (built on TORCS) and Flow \cite{wu2021flow} (built on SUMO) are frequently employed. These tools facilitate the study of interactions and behaviors in complex traffic and multi-agent environments. In Table \ref{tab:platforms_sw}, a summary of simulators used for main small-scale car platforms is shown.\par  

Communication in robotics is as critical as the simulator itself, and ROS \cite{quigley2009ros} serves as a key middleware platform for managing robotic software development. ROS facilitates communication between various robotic components, including sensors, actuators, and decision-making algorithms, enabling seamless integration and scalability. It also offers a vast collection of pre-built packages and libraries for common tasks like motion planning and image processing. Currently, ROS exists in two versions: ROS 1 and ROS 2. ROS 1, the original version, is widely used and supported by the community; however, its centralized architecture limits real-time capabilities and lacks robust security features. In contrast, ROS 2 addresses these limitations with a decentralized design based on the Data Distribution Service, offering enhanced scalability, real-time support, and Quality of Service. While ROS 1 remains suitable for projects reliant on mature tools and is ideal for educational or non-real-time applications, ROS 2 is recommended for new projects requiring real-time performance, high scalability, security, and cross-platform compatibility. Nevertheless, the Simulation to Reality (Sim2Real) gap remains a challenge that needs to be addressed before successfully deploying the trained algorithms in the real world.

\subsection{Sim2Real Transfer}

Sim2Real is a concept in robotics and AVs that involves transferring skills, knowledge, or models acquired in a simulated environment to real-world applications. Here, the real-world setting or scenario in which the robot is intended to operate and execute tasks is termed the target domain. Conversely, the data, and experiences, shaping the development of the robotic system are called the source domain. The core objective of Sim2Real is to devise algorithms and methodologies capable of effectively bridging the disparities between these two domains, known as the \textit{Sim2Real gap}. For instance, this gap may manifest as dynamic differences or discrepancies in the sensing part, where the simulated images and the real images are different. Although the Sim2Real transfer is primarily centered on transferring Deep Reinforcement Learning (DRL) policies from simulation to the real world, it can also be more broadly considered as ML problems for the sensing part, for the agent facing situations in the real world that have not appeared in simulation. To address the Sim2Real gap, an array of methods is proposed, including system modeling, dynamics randomization, and randomization for sensing. Here we introduce the most used methods for small-scale cars.\par

The first approach for the real-world application of small-scale cars is \emph{Zero-shot Transfer}, where the trained model is directly applied in real-world settings. Imitation Learning (IL), based on datasets from real platforms, trains agents to mimic expert behavior \cite{rosman2017hybrid, dey2021joint, goldfain2019autorally, drews2017aggressive, pan2017agile, drews2019vision, stocco2022mind, kannapiran2020go}. While efficient, this method assumes simulation and real-world environments are similar. DRL-based Zero-Shot Transfer, achieved via compact observation and output spaces \cite{tan2018sim, li2024platform, li2023vision}, depends heavily on simulation fidelity for success.  Another noteworthy approach is \emph{Transfer Learning}, which aims to improve the performance of the target agent in the target domain by transferring the knowledge contained in different but related source domains \cite{zhuang2020comprehensive}. It excels when simulation closely aligns with real-world scenarios but falters with significant domain mismatches. For small-scale cars, this method balances efficiency and adaptability in controlled conditions. When the labeled data in the target domain is scarce or expensive, \emph{Domain Adaptation} is used. As a subset of transfer learning, it seeks specifically to minimize the distribution mismatch between the source and target domains, allowing the model to generalize better across different domains. It is effective in dynamic environments but requires computational resources and high-quality feature extraction, posing challenges in real-time applications. Another frequently used approach is \emph{Domain Randomization}, in which the randomness and variation are introduced during simulation to make the model more robust to different real-world conditions \cite{tobin2017domain}. It works by for example randomly varying simulation parameters, such as lighting conditions, textures, object appearances, and physics properties during training. For small-scale car platforms, Duckietown and DeepRacer provide \emph{Domain Randomization} option in the simulation, which is easier for the users to tackle Sim2Real issue \cite{candela2022transferring, balaji2020deepracer, kalapos2020sim, sandha2021sim2real}. While fostering generalization, excessive randomization may dilute the relevance of training data, compromising real-world performance.

\section{Sensors and Sensor systems}
\label{sec:sensors}

In Section \ref{subsec:platform}, various platforms are explored, each employing diverse sensors to perceive their surroundings and execute tasks. Compared with full-scale AVs, which are equipped with various state-of-the-art sensors for perception \cite{ignatious2022overview}, small-scale cars usually use lightweight sensors. Here, we discuss the most commonly utilized sensors across these platforms, providing their respective use cases.

\subsection{Camera}
\label{subsec:camera}
An RGB camera provides a stream of images that is easily processed and understood by humans. This helps researchers to better understand the perspective of the robot. The amount of information provided by cameras is huge, which allows the robot to perceive a lot of information with just one sensor in a short amount of time. The downside of this is, that they need a high bandwidth communication link to whichever part of the robot needs those information. Also with the information in images being so densely encoded, it is difficult for a robot to understand the environment.\par

The field of computer vision has emerged as a crucial scientific discipline, facilitating the extraction of pertinent details from images and videos. In recent years, the advent of deep learning (DL) models, such as You Only Look Once (YOLO) \cite{redmon2016you}, has significantly contributed to enhancing the robot's ability to interpret scenes captured by cameras. It is worth noting that cameras face external influences, such as varying lighting conditions, which can substantially alter the captured images. In extreme scenarios, such as very dark environments or instances where direct light affects the camera, the effectiveness of cameras may be compromised.
A big advantage of cameras is their wide pricing range, starting from 15 USD, corresponding to images of different quality in regards to resolution and dynamic range. They are also easy to interface with, usually requiring only a USB connection. Some tasks can be accomplished only with cameras, e.g. traffic sign or road marking detection  \cite{satilmics2019cnn, zannatha2022integration, asghar2023control}.\par

Beyond RGB cameras, more advanced options exist, including those detecting optical flow \cite{shiba2022secrets} (the apparent motion of objects) or depth cameras \cite{lee2012depth}, employing different methods to ascertain the distance to objects. These advanced cameras provide both depth information and standard RGB images.
In the context of autonomous tasks for single-car platforms discussed in this paper, cameras emerge as a ubiquitous choice. Their cost-effectiveness and information-rich output make them the preferred sensor for many platforms. Among the commonly used cameras in scaled vehicles, the Raspberry Pi Camera stands out \cite{paull2017duckietown, donkeycar, wilson2016pheeno, kannapiran2020go, pohlmann2022ros2, samak2023autodrive}. Designed for the Raspberry Pi, it combines affordability with decent image quality, offering a resolution of 1080p and a dynamic range of 44dB. Another popular option used in the discussed platforms is the Intel Realsense line of cameras \cite{srinivasa2019mushr, vincke2021open}. Those include two cameras and an IR projector, which allow to capture not only an RGB image but also depth information of each pixel using stereovision. The IR projector helps to improve accuracy in scenes with poor textures. The manufacturer claims an accuracy in the centimeter range.

The big advantage of cameras is their versatility combined with their low price. Their dense stream of information makes them suitable for almost all tasks discussed in this paper. Although different approaches for processing the information from cameras need to be applied depending on the task. Applying DL approaches helps with this in many different fields and allows for quickly iterating through different approaches. One major drawback of relying solely on cameras is their susceptibility to varying environmental conditions, as previously mentioned. This sensitivity complicates the transfer of control strategies from simulations to real-world vehicles. For successful transfer, simulations must either model these environmental influences with high accuracy or incorporate an additional interpretation layer that bridges the gap between simulation and reality.

\subsection{LiDAR}
\label{subsec:lidar}
LiDAR, another widely employed sensor in numerous platforms, functions by measuring the time taken for a beam of light to reflect off a surface. The LiDAR emits beams from multiple angles—typically ranging from 360 to 2048—creating a detailed point cloud map of objects in the environment. Multi-layer LiDAR systems, with laser beams spanning up to 128 angles, provide data on objects at different heights, generating a comprehensive 3D point cloud. LiDARs are available in various resolutions, offering enhanced detail, especially in complex and structured environments. The typical scanning frequency of LiDARs ranges from 5 to 20 Hz, with high-end models achieving detection ranges of up to 350 meters.\par

The distinct advantage of LiDAR over cameras lies in its direct identification of the surroundings, including the precise position of objects relative to the robot. In contrast, a camera first needs to identify an object in an image and then estimate its position. Moreover, LiDAR provides significantly more accurate positional information compared to estimations derived from camera images. However, unlike cameras, LiDARs lack the ability to discern details such as the color of the detected objects. This limitation renders them unsuitable, for example, in tasks like traffic signs or lane detection. LiDAR also has a significant advantage over cameras in that it is unaffected by environmental conditions such as varying lighting, and it can operate effectively in complete darkness. Additionally, some LiDAR models offer reflectivity data, which can help distinguish between different surface types. For example, they can differentiate highly reflective road markings from less reflective obstacles. However, LiDARs are typically more expensive than cameras. The most affordable hobbyist LiDARs start at around 150 USD for a basic single-plane sensor, with prices increasing rapidly as resolution or the number of planes expands.\par

Usually physically larger platforms use LiDAR sensors \cite{balaji2020deepracer, donkeycar, o2020f1tenth, karaman2017project, gonzales2016autonomous, srinivasa2019mushr, vincke2021open, kannapiran2020go, berlin2020autominy, samak2023autodrive, elmquist2022software} as they tend to be bigger and heavier than cameras. Those platforms are also usually more versatile and used for more different tasks and therefore the additional cost of a LiDAR is feasible. In our investigation, all platforms utilizing LiDAR also integrate a camera to address the limitations of LiDAR for specific tasks. The LIDAR that is used by far the most on the platforms we investigated is the single plane RPLIDAR \cite{donkeycar, gonzales2016autonomous, vincke2021open, kannapiran2020go, berlin2020autominy, samak2023autodrive}. It offers a reasonable detection range of 12m and a resolution of 0.225 degrees for small-scale car platforms with a 360 field of view. And is one of the cheaper options costing around 400 USD. But it only allows for a 10 Hz scan frequency. The second most popular, higher end solution is the Hokuyo UST-10LX \cite{o2020f1tenth, karaman2017project}, offering a 270 degree field of view and a 10m detection range. 

Particularly tasks, that require precise information about the location and orientation of surrounding vehicles or obstacles benefit greatly from the improved accuracy of LIDARs over cameras, these include car following, overtaking, or obstacle avoidance. Tasks such as lane following or traffic sign following need the additional visual information provided by cameras and can therefore not be accomplished with only LIDARs.

For economic considerations, cost-effective sensors are the mainstream choice for small-scale cars. These sensors typically offer lower resolution but demand less computational power. For example, the RPLiDAR is commonly used in small-scale cars, providing a detection range of up to 40 meters with error margins of approximately 1\%. It has a data rate of around 32,000 samples per second. In contrast, typical LiDAR sensors used in full-scale AVs offer significantly higher range, resolution, and data rates. For instance, the Velodyne LiDAR sensor can detect up to 120 meters with a 2 cm error margin and a data rate of 1.3 million points per second. These hardware limitations make it challenging to develop new approaches, leading primarily to the application of mature AV techniques from full-scale cars under simplified conditions. However, advancements in semiconductor technology are expected to make more sophisticated hardware setups available for small-scale cars in the near future, which will unlock more advanced AV techniques for small-scale cars.

\subsection{IMU and Encoder}
Some sensors that are usually not used on their own but as an additional source of information for other sensors are IMU and odometry encoders.
Where odometry encoders provide information about the angle by which each wheel of a robot has turned and therefore allow us to estimate where it has traveled, the IMU provides information on how fast a robot is spinning along each of its axes aiding in estimating the total rotation.\par

The information of these two types of sensors is usually combined with other absolute positioning inputs like a camera or a central localization system to fuse them in a filter like a Kalman filter \cite{kalman1960new} or in any other kind of perception module. IMUs are available at relatively affordable prices, ranging from 15 USD. The two IMU sensors most commonly used in the discussed platforms are the MPU9250 and MPU6050 \cite{paull2017duckietown, donkeycar, mondada2009puck, vincke2021open, samak2023autodrive}. Although it has to be noted, that not all publications mention the exact model of the IMU used. Both sensors feature a 3-axis gyroscope, which measures the rotational speed, and a 3-axis accelerometer, which measures linear acceleration. The MPU9250 additionally offers a 3-axis magnetometer to incorporate measurements of the Earth's magnetic field into orientation sensing.\par

Odometry encoders can be found in several different forms, ranging from magnetic encoders to optical ones. Their implementation is usually dependent on the platform, as they need to be tightly integrated into the mechanics of the robot. Despite their low cost, often under 10 USD, they offer reliable positional data, making them a popular choice. Almost all the platforms reviewed in this paper incorporate an IMU and some form of encoder, as these sensors are both affordable and effective, particularly in slow, controlled indoor environments. The notable exception is the smallest platform, primarily designed for swarm robotics research \cite{rubenstein2012kilobot}, where extreme cost constraints led to the exclusion of these sensors.

The addition of cheap odometry encoders or IMUs especially benefits tasks that require the knowledge of the precise location and orientation of the vehicles. This includes tasks such as path following, racing, or drifting. The disadvantage of this type of sensor lies in the need to incorporate it in a meaningful way, this usually involves either existing filters or using deep neural networks. In the first case, the additional filter needs to be tuned to provide meaningful information, in the case of the latter the training of the neural network becomes more complex. The big advantage lies in the small price and easily accessible information.

\subsection{GPS}
GPS is the least frequently used sensor with model scale platforms \cite{goldfain2019autorally, gonzales2016autonomous} offer positional information in a global coordinate system. Most of the discussed platforms are designed for indoor use, which eliminates the use of GPS as the signal does not penetrate buildings. However, recognizing the pivotal role GPS plays in full-sized vehicles, many model-scale platforms opt to replace it with indoor central localization systems, as detailed in Section \ref{subsec:central_loc}, to enhance realism. GPS receivers are also comparatively cheap at around 30 USD. However, for those seeking higher precision, the cost can escalate to several hundred USD, especially for the more advanced GPS RTK systems \cite{feng2008gps}. Such precision may become necessary for model-scale platforms, as conventional GPS systems typically only offer an accuracy of approximately 5 meters \cite{team2014global}.
The biggest disadvantage of GPS sensors for small-scale vehicles is their reliance on outdoor environments, which require significantly more space and make experiments heavily dependent on weather and other outdoor conditions. As a result, GPS sensors are only beneficial for platforms when the research focuses specifically on the accurate effects and phenomena associated with GPS data, such as multipath errors or signal interference.

\subsection{Central Localization System}
\label{subsec:central_loc}
In situations where the emphasis is on studying the interaction among multiple vehicles rather than individual vehicle behavior, a central localization system streamlines the task, allowing a more focused approach to other aspects such as obstacle avoidance or overtaking, by providing precise information about the position and orientation of different objects and sometimes even their velocity. By decoupling the algorithms for (multi) vehicle behavior from the perception of individual vehicles, central systems enable the study of system dynamics in an idealized "best case" scenario. In this context, the term "best case" refers to a situation where a vehicle possesses precise knowledge of its own position and that of surrounding obstacles. However, it is essential to acknowledge that this assumption may not align with the complexities of real-world road situations. Conversely, central systems find utility in simulating GPS positioning within actual vehicles.\par

One of the cheapest options implemented in the considered platforms involves the use of special markers or tags on each vehicle, which are observed by one or multiple overhead cameras \cite{zdevsar2022cyber}. These markers, based on ArUco \cite{aruco2014} technology, are strategically placed on vehicles and key positions or obstacles. Knowing the intrinsic and extrinsic parameters of the cameras, existing libraries like OpenCV \cite{bradski2000opencv} can be employed to detect these markers and estimate their position and orientation. As discussed in Section \ref{subsec:camera}, cameras are relatively inexpensive, and the markers merely need to be printed out. While the resulting localization precision is reasonable \cite{xavier2017accuracy}, it does decline with the distance of the markers from the camera. Mitigating this issue involves the use of multiple cameras, necessitating synchronization and precise positioning of all cameras.\par

Off-the-shelf solutions for tracking also exist from various manufacturers, such as OptiTrack or VICON. These solutions predominantly rely on high-speed IR cameras and reflective markers on the tracked objects. The systems inherently provide position and orientation information for different marker assemblies (referred to as rigid bodies) without requiring additional processing. Notably, the accuracy of these systems surpasses that of solutions utilizing RGB cameras, with manufacturers often specifying sub-millimeter accuracy. However, the upfront cost is relatively high, typically exceeding 10,000 USD, and in some cases, additional licensing fees for the requisite software may apply \cite{hyldmar2019fleet, stager2018scaled}.

Central localization systems offer the advantage of high precision localization of any object within the test setup and are easily extendable to more vehicles of objects as no additional sensors need to be mounted on the vehicles themselves. They are however quite expensive and require a dedicated mounting setup to position the sensors, therefore they cannot easily be relocated. Their high precision allows for easy transferability from simulation to real-world testbed. The implementation of these systems especially benefits tasks, that do not study the influence of sensors and sensor behavior on the control algorithm, but merely the control strategies themselves. They can essentially be used as the single source of information for any of the tasks listed except traffic sign following.

\subsection{Other sensors}
Several platforms discussed in Section \ref{subsec:platform} employ additional sensors to either emulate sensors found in full-scale autonomous cars or fulfill specific tasks.\par

A prevalent type of sensor utilized in many platforms \cite{paull2017duckietown, mondada2009puck, riedo2013thymio, bonani2010marxbot, vincke2021open, kannapiran2020go, pickem2015gritsbot, betthauser2014wolfbot, pohlmann2022ros2, rezgui2020autonomous} is the single-point distance sensor, available as either ultrasonic or infrared. These sensors operate on the time-of-flight principle. They measure the time it takes for a wave to reflect back to the sensor and calculate the distance accordingly. Typically positioned at the front of vehicles, they often mimic radar sensors and play a crucial role in tasks such as car following. \par

Line following sensors, consisting of an array of light sensors aimed at the ground, are employed in certain platforms \cite{vincke2021open, stager2018scaled, vincke2021open} to assist visual line following, with a camera serving as a secondary input. These sensors can easily discern the line from the road by measuring the reflectance of the surface on multiple points along the length of the sensor. A limitation is their coverage area, which is relatively small and located underneath the car. If the vehicle deviates significantly from the line, the sensor may struggle to rediscover it.\par

One platform \cite{zdevsar2022cyber} utilizes RFID readers to detect specific points of interest on the road, such as the beginning of intersections or as indicators for turns on a crossroad. Turtlebot employs cliff sensors, which function as downward-facing distance sensors, indicating if the part of the vehicle with the sensor is suspended over a steep drop. This sensor serves to prevent the vehicle from inadvertently falling off a precipice in the environment. \par

For the platform designed for use with a high amount of robots \cite{rubenstein2012kilobot}, infrared (IR) transmitters and receivers are employed. These facilitate cost-effective communication between different robots and emulate a simple form of Vehicle-to-Vehicle (V2V) communication.

\subsection{Compute Units}

To process the information from the sensors and compute actuator movement according to their task, the platforms require a dedicated computational unit. As the physical size of the platforms is quite limited, the computational performance of these compute units is as well. The prevailing choice across numerous platforms \cite{paull2017duckietown, donkeycar, vincke2021open,  stager2018scaled, hyldmar2019fleet, morrissett2019physical, kloock2021cyber, zdevsar2022cyber, pohlmann2022ros2} is the Raspberry Pi single-board computer (SBC). It offers a performance of about 32 GFLOPS leveraging both CPU and GPU. Built on a System on a Chip (SoC), the Raspberry Pi features a compact physical size but offers modest computational performance, with a 4-core 1.5GHz CPU. Despite its relatively low processing power, the Raspberry Pi is well-suited for the majority of tasks performed by these platforms. The Raspberry Pi also benefits from a large developer and research community providing a diverse set of frameworks and libraries to harness its capabilities.\par

A comparable alternative to the Raspberry Pi is the Nvidia Jetson. Also based on an SoC, the Nvidia Jetson stands out by incorporating an additional higher performance GPU to accelerate ML tasks, resulting in a performance of around 472 GFLOPS. Different versions of the Nvidia Jetson are available, some also offering dedicated resources for tensor operations for DL tasks. The highest performing option offers up to 300 TOPS. Given their similar connectivity, some platforms offer compatibility with either the Raspberry Pi or the Nvidia Jetson \cite{paull2017duckietown, donkeycar}. However, the majority of platforms exclusively support the Nvidia Jetson \cite{o2020f1tenth, karaman2017project, srinivasa2019mushr, elmquist2022software, samak2023autodrive, wang2019hydraone}.\par

For platforms not requiring high computational performance on the vehicles themselves, microcontrollers \cite{mondada2009puck, riedo2013thymio, carron2023chronos, rubenstein2012kilobot, robinette2009labrat, pickem2015gritsbot} are very common. Although microcontrollers offer lower performance than Raspberry Pi, offering only between 100 KFLOPS and 25 MFLOPS, they enable real-time code execution, crucial for interfacing with low-level sensors and actuators. They also use less power and are cheaper. Conversely, some platforms leverage regular PCs as their computational unit \cite{balaji2020deepracer, goldfain2019autorally}. While PCs provide the highest performance, ranging from 50 GFLOPS for CPU-only tasks up to multiple TFLOPS for GPU computation, they have larger physical footprints and higher power consumption. Because of this, some platforms \cite{pohlmann2022ros2, stager2018scaled, hyldmar2019fleet} opt to use a lower power compute unit like a Raspberry Pi on the vehicles themselves for command execution and sensor data acquisition only and offload compute-intensive tasks such as control strategies and computer vision algorithms to a more powerful central computer that is connected through a network to the compute units on the vehicles. This increases the complexity of the setup as network communication has to be handled but allows for easier reconfiguration and upgrades as only the central compute unit has to be changed. This also allows to use external high performance compute clusters to handle complex control strategies.\par

Especially control strategies involving a camera as a sensor or strategies fusing multiple sensors require significant performance from the computing unit. Platforms that employ these strategies usually use one of the higher end options listed, namely an Nvidia Jetson or regular PCs, either on the vehicles or as a central computer unit. The smaller and the cheaper the platforms tend to be, the smaller also the compute units have to be, as size and energy constraints increase with smaller sizes. Smaller platforms also usually employ smaller and more low level sensors which can only be interfaced with through low level compute units like microcontrollers or SBCs.

\subsection{Vehicle-to-Everything (V2X)}

V2V refers to the ability of autonomous vehicles to communicate with one another to collaboratively perform cooperative driving tasks. Its extension, Vehicle-to-Infrastructure (V2I), involves communication between vehicles and infrastructure elements, such as traffic lights or traffic signs. In 2019, the definition was expanded to V2X, which encompasses V2V and V2I, but also extends to communication with broader systems like the power grid or cellular networks. In full-scale autonomous vehicles, dedicated network components and protocols facilitate this communication.\par

Small-scale vehicles play a crucial role in researching these types of interactions, as they enable easy deployment of complex setups involving both vehicles and infrastructure in compact testbeds.  For a small-scale platform to support such research, it must have networking capabilities, which excludes many of the educational platforms discussed in Section \ref{subsubsec:educational_platforms}. Most other platforms rely on WiFi for networking, with the notable exception of the Kilobot \cite{rubenstein2012kilobot}, which uses a proprietary infrared communication link. WiFi enables both centralized approaches, where all vehicles and infrastructure elements communicate with a central coordinator, and decentralized approaches, where vehicles communicate directly with each other.  All of the platforms investigated in our research rely on a central coordinator to manage communication, as seen in \cite{stager2018scaled} and \cite{hyldmar2019fleet}. However, the open-source platforms referenced in Section \ref{subsubsec:opensource_platforms} can be extended to support alternative communication strategies. Most of the platforms we examined only support V2V communication, with the Duckietown \cite{duckietown} and AutoDRIVE \cite{samak2023autodrive} platforms being exceptions. These two platforms also support V2I communication in the form of connected traffic lights. In Duckietown, the traffic lights are essentially built with the same hardware as the vehicles, allowing them to sense other vehicles through onboard cameras and accept control commands from either a central coordinator or other vehicles.

\begin{figure*}
    \centering
    \includegraphics[width=0.85\textwidth]{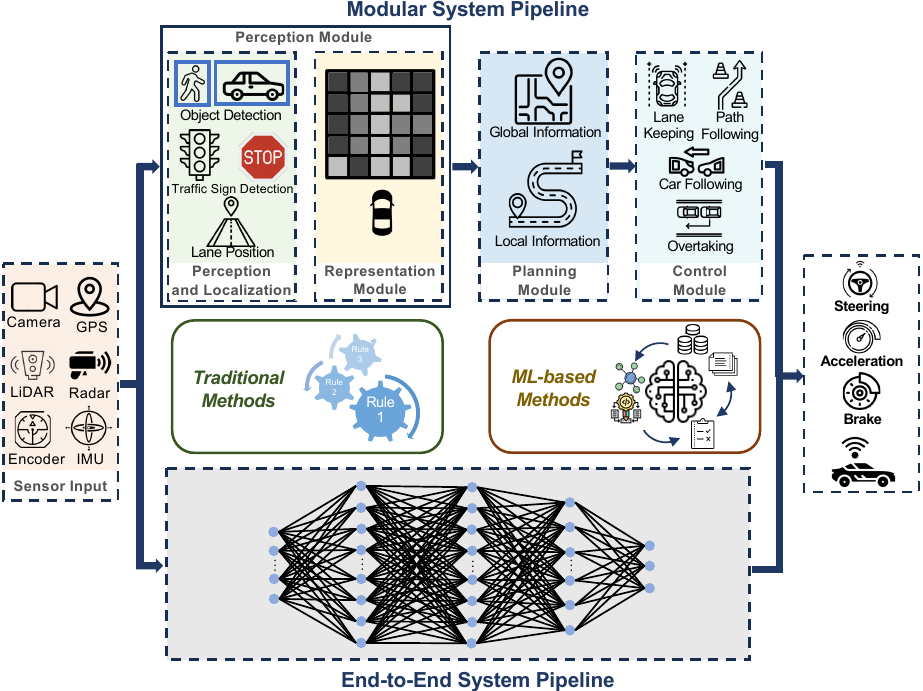}
    \caption{Comparison of two pipelines for the autonomous driving system. An end-to-end system maps raw sensor inputs directly into control commands, whereas a modular system includes multiple subsystems to process the sensor inputs sequentially and output control commands.}
    \label{fig:system compare}
\end{figure*}

\section{Benchmarking}
\label{sec:Task}

As discussed in Section \ref{sec:sensors}, small-scale cars primarily apply mature AV techniques from full-scale cars under simplified conditions. Thus, autonomous driving systems for small-scale vehicles are typically categorized into two distinct pipelines, similar to those used in full-scale cars: the \emph{end-to-end system} pipeline and the \emph{modular system} pipeline. As illustrated in Fig. \ref{fig:system compare}, a modular system comprises multiple subsystems that perform various tasks such as perception and localization, mapping, path planning, and control \cite{grigorescu2020survey, kiran2021deep}. Each subsystem focuses on specific functionalities and tasks. First is the perception system, where sensors such as cameras and LiDAR are used to gather information about the surroundings, including lane markings, obstacles, traffic signs, and other vehicles. Following this, the localization and mapping system leverages GPS, odometry, IMU, or techniques like simultaneous localization and mapping (SLAM) to precisely pinpoint the position within the environment while concurrently creating a detailed map of its surroundings. The perception system creates an intermediate representation of the environment for subsequent utilization. \par

The representation module then uses the information from the perception module and further processes the sensor data with sensor fusion techniques or creates an object map with the predicted state of each object within the sensor range. The combination of the perception module and the representation module can be seen as the scene understanding, which provides an abstract high-level representation of the environment \cite{kiran2021deep}. Afterward, the planning system maps out a safe and efficient route to reach the destination. Normally, in the autonomous driving system, the planning phase is divided into two different parts, namely \textit{global path planning} and \textit{local path planning} \cite{teng2023motion}. Global path planning refers to the process of determining an optimal or feasible route from the current position to its destination and is done considering the entire environment and involves high-level decision-making. Local path planning or path following, on the other hand, focuses on the immediate surroundings of the vehicle and deals with making real-time adjustments to adhere to the planned global path. Finally, the control module generates driving commands based on the processed information.  In control theory, the primary goal is to minimize a cost function. Various methodologies are employed for the control system, classical controllers such as the Proportional–Integral–Derivative (PID) controller, Model Predictive Control (MPC), and ML-based controllers such as IL or DRL.  PID works by continuously calculating the error between a desired setpoint and a measured process variable, then applying corrective actions based on three components: proportional, integral, and derivative. The proportional component reacts to the current error, providing immediate corrections, the integral component addresses accumulated past errors to eliminate steady-state error, and the derivative component predicts future errors by considering the rate of change, enhancing stability and responsiveness. MPC, on the other hand, is an advanced control strategy designed for complex systems. It predicts future system behavior using a dynamic model and computes optimal control inputs by solving a constrained optimization problem at every time step.

\begin{figure*}
    \centering
    \includegraphics[width=0.6\textwidth]{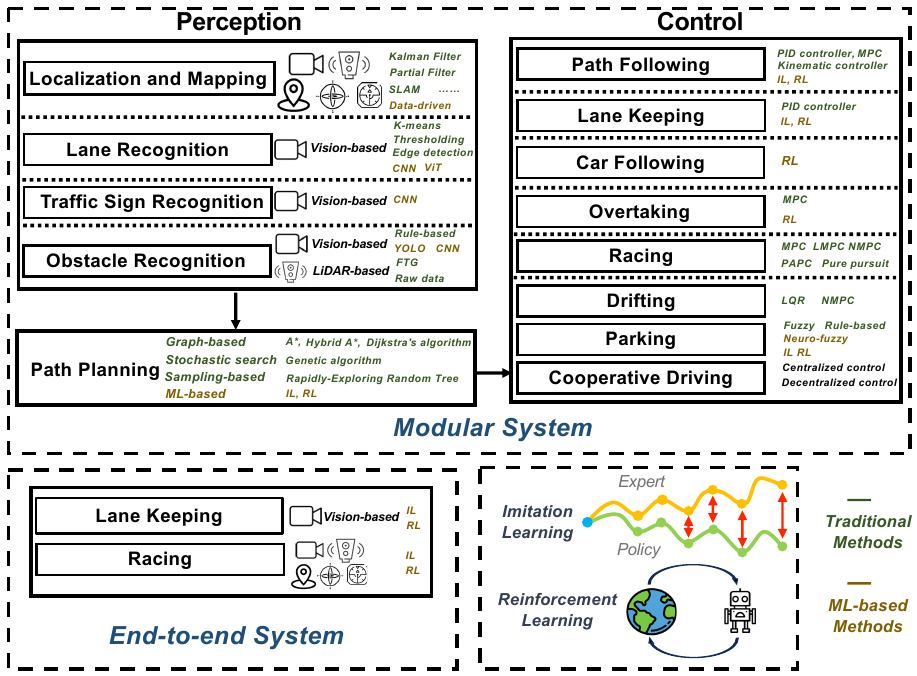}
    \caption{Classification of research in small-scale cars into modular and end-to-end systems. Modular systems are further subdivided into perception tasks, planning tasks, and control tasks.}
    \label{fig:structure of task}
\end{figure*}

For ML-based controllers, as illustrated in Fig. \ref{fig:structure of task}, IL involves training an agent to mimic expert behavior by learning from a dataset of demonstrations. By directly leveraging expert trajectories, IL bypasses the need for explicitly defined reward functions, making it particularly effective in scenarios where optimal behavior is well understood but challenging to formalize mathematically.  Key IL algorithms include Behavioral Cloning (BC) \cite{bain1995framework}, Generative Adversarial Imitation Learning (GAIL) \cite{ho2016generative}, Inverse Reinforcement Learning (IRL) \cite{ng2000algorithms}, and Dataset Aggregation (DAgger) \cite{ross2011reduction}. On the other hand, RL involves training an agent to autonomously discover optimal policies by interacting with an environment and receiving feedback in the form of rewards \cite{sutton2018reinforcement}. This exploration-based approach allows RL agents to learn from scratch and adapt to a wide range of environments, even in the absence of expert guidance. \par

An end-to-end system is characterized by a unified architecture that aims to learn the entire mapping directly from raw sensor inputs to driving actions without explicitly decomposing the task into separate modules \cite{tampuu2020survey, le2022survey, chib2023recent}. It involves training a comprehensive learning model with ML methods, IL or DRL, directly processing raw sensor data and output control commands. End-to-end systems potentially simplify the system architecture by eliminating the need for handcrafted modules and feature engineering.\par

In this section, we focus on the two primary systems and the key tasks associated with each, as depicted in Fig. \ref{fig:structure of task}. Rather than delving into the detailed methodologies used to accomplish these tasks, we provide references to the relevant studies for further exploration. Additionally, we propose a baseline framework for each system, aiming to provide inspiration and guidance for researchers in developing their own approaches.

\subsection{Perception}

In a modular system, the perception system performs several key functions essential to autonomous driving, including mapping the environment and localizing the autonomous vehicle, detecting lanes, and detecting objects. In this subsection, we will discuss various aspects that the perception module needs to consider.

\subsubsection{Localization and Mapping}
\label{subsec: localization}

Localization and mapping are foundational tasks in autonomous systems, often requiring high precision and real-time performance. A primary consideration in localization is GPS, which offers direct positional data for the vehicle. However, the accuracy of GPS is influenced by various factors, such as atmospheric conditions, satellite geometry, signal blockage (buildings, trees, etc.), multipath interference, and the quality of the GPS receiver. Differential GPS (DGPS) systems and augmentation techniques can be used to enhance GPS accuracy by correcting some of these error sources, but cannot fulfill the accuracy which is extremely important when it is small-scale cars. IMUs, another commonly used sensor, provide high-rate measurements of acceleration and angular velocity across three dimensions but suffer from issues like sensor drift and integration errors, leading to long-term inaccuracies. To address the issues, the two sensors are often integrated through sensor fusion methods, such as Kalman filter (KF) \cite{kalman1960new} to enhance accuracy and reliability in determining the position and orientation. As in the studies by \cite{brunner2017repetitive, rosolia2019simple, alcala2020autonomous}, that use BARC small-scale cars, KF makes the fusion of IMU, wheel encoders, and indoor GPS measurements to achieve an accurate localization. Nevertheless, KFs are limited by assumptions of linearity and Gaussian noise, making them less effective in complex or nonlinear environments. Therefore, \cite{goldfain2019autorally} and \cite{williams2016aggressive} use factor graphs combined with incremental smoothing and mapping 2 (iSAM2) \cite{kaess2012isam2}, to fuse GPS and IMU measurements and output smoothing estimation. For the issues with non-linear dynamics and non-linear measurement models, \cite{drews2019vision, lee2020perceptual, wagener2019online} use the particle filter to fuse the camera images with IMU, and GPS. \cite{liu2021navigation} also presents an approach to use KF to fuse multiple-camera images with the AprilTag system and wheel odometry, validated using the OptiTrack Motion Capture System, which offers high accuracy but at a significantly higher cost, limiting its practicality for small-scale platforms. Additionally, a single camera can serve as a perception sensor. For example, in \cite{drews2017aggressive}, CNN is used with a single monocular camera to predict the cost map of the track in front of the vehicle which is directly useable for online trajectory optimization with MPC. Similarly, \cite{zheng2019vision} uses a top-down lane cost map CNN and the YOLOv2 CNN to extract feature-input values and a two-point visual driver control model (TPVDCM) as the controller to control the vehicle. However, these camera-based perception approaches rely solely on visual information, neglecting spatial information, which can lead to limitations in accuracy and robustness under certain conditions, where the camera is not robust enough. \par

LiDAR-based systems have emerged as a compelling alternative, leveraging 2D LiDAR sensors for localization and mapping via SLAM techniques. A well-established method for localizing robot cars with LiDAR sensors on the pre-defined map involves the use of a particle filter (PF) as explored in \cite{sinha2020formulazero} with F1TENTH. However, the computational demands of the PF present challenges, especially for computation resources and space-constrained small-scale robot cars. To enhance performance, \cite{walsh2018cddt} introduces the Compressed Directional Distance Transform (CDDT) on the RACECAR platform. This method focuses on expediting ray casting within 2D occupancy grid maps and accelerating sensor model computations to mitigate computational expenses. While comprehensive survey papers delve into various SLAM techniques \cite{takleh2018brief, kazerouni2022survey, khan2021comparative, alsadik2021simultaneous}, encompassing feature-based, LiDAR-based, visual, and graph SLAM, this study offers a concise overview of three prevalent SLAM methods employed with small-scale cars: GMapping  \cite{sankalprajan2020comparative, ngo2020towards, liu2022research}, HectorSLAM \cite{nobis2019autonomous, massa2020lidar, abdelrasoul2016quantitative, zhao2022research, post2017autonomous}, and Cartographer  \cite{zhu2019design, filipenko2018comparison, zhang20202d}. While 3D LiDAR remains underutilized due to its size and computational demands, advancements in hardware miniaturization are making 3D LiDAR SLAM techniques like LOAM \cite{zhang2014loam} and LIO-SAM \cite{shan2020lio} increasingly feasible for small-scale platforms.\par

Another compelling approach gaining traction involves leveraging ML methods to address mapping and localization challenges. One notable instance is the GALNet proposed in \cite{mendoza2020galnet} implemented on the Autominy platform. GALNet employs a Deep Neural Network (DNN), utilizing two timestamps of inertial, kinematic, and wheel velocity data to estimate poses effectively. With the success of Transformer architecture in various research fields, \cite{bonatti2023pact} introduces the Perception-Action Causal Transformer (PACT) architecture. This model constructs a representation from sensor data by autonomously predicting states and actions over time, laying the groundwork for subsequent task-specific networks for localization and mapping. Pre-trained models are employed initially and later fine-tuned for specific tasks, demonstrating validation on MuSHR with LiDAR sensor data. For image-based localization, \cite{garcia2023deep} proposes an effective framework leveraging a pre-trained local feature transformer (LoFTR). This framework employs a constrained 3D projective transformation between consecutive key images to establish a visual map on Jetbot. These studies underscore the growing interest and potential of ML methodologies, including neural networks and transformer architectures, in revolutionizing mapping and localization paradigms. However, ML approaches are often data-intensive and require extensive pre-training and fine-tuning to generalize effectively. Future research should focus on improving the robustness and adaptability of these methods, integrating advanced sensing technologies, and optimizing computational efficiency to bridge the gap between traditional and data-driven localization paradigms.

\begin{table}
    \centering
    \scriptsize
    \caption{Comparison of methods used for localization and mapping.}
    \label{tab:my_label}
    \resizebox{\linewidth}{!}{%
    \begin{tabular}{|l|c|l|l|}
    \hline
           \textbf{Method} & \textbf{Sensor} & \textbf{Analysis} & \textbf{Reference} \\
    \hline
         \textbf{Kalman Filter} & \makecell[l]{GPS; IMU; \\Encoder}&  \makecell[l]{ $ \textcolor{green}{\raisebox{0.6ex}{\scalebox{0.7}{$\sqrt{}$}}}$  Accuracy improvement \\ $ \textcolor{red}{\scalebox{0.85}[1]{$\times$}}$ Asynchrony, Latency \\ \textcolor{red}{\scalebox{0.85}[1]{$\times$}} Noise Assumptions } 
         & \makecell[l]{ \cite{brunner2017repetitive, rosolia2019simple, alcala2020autonomous, xue2023learning}: BARC; \\ \cite{liu2021navigation}: QCar} \\
         
    \hline
         \textbf{Particle Filter} & \makecell[l]{GPS; IMU; \\ LiDAR}   &  
         \makecell[l]{ $ \textcolor{green}{\raisebox{0.6ex}{\scalebox{0.7}{$\sqrt{}$}}}$  Nonlinear dynamics \\ $ \textcolor{red}{\scalebox{0.85}[1]{$\times$}}$ High computational demand} 
         & \makecell[l]{ \cite{sinha2020formulazero}: F1TENTH; \cite{walsh2018cddt}: RACECAR} \\
         
    \hline
         \textbf{GMapping} & \makecell[l]{LiDAR}
         &  \makecell[l]{ $ \textcolor{green}{\raisebox{0.6ex}{\scalebox{0.7}{$\sqrt{}$}}}$ Robust in static environments \\ $ \textcolor{red}{\scalebox{0.85}[1]{$\times$}}$ High memory usage} 
         & \makecell[l]{ \cite{nobis2019autonomous}: Roborace; \cite{massa2020lidar,zhao2022research}: - }\\

    \hline
         \textbf{HectorSLAM} & \makecell[l]{LiDAR}
         & \makecell[l]{ $ \textcolor{green}{\raisebox{0.6ex}{\scalebox{0.7}{$\sqrt{}$}}}$  Fast and lightweight \\ $ \textcolor{red}{\scalebox{0.85}[1]{$\times$}}$ Performance degrades with noise} 
         & \makecell[l]{ \cite{sankalprajan2020comparative,ngo2020towards,liu2022research}: -}\\

    \hline
         \textbf{Cartographer} & \makecell[l]{LiDAR}
         & \makecell[l]{ $ \textcolor{green}{\raisebox{0.6ex}{\scalebox{0.7}{$\sqrt{}$}}}$ Accuracy, dynamic environments \\ $ \textcolor{red}{\scalebox{0.85}[1]{$\times$}}$ High computational demand} 
         & \makecell[l]{  \cite{klapalek2021comparison}: F1TENTH; \cite{filipenko2018comparison}: Innopolis UGV; \\ \cite{zhu2019design, zhang20202d}: -}\\
         
    \hline
         \textbf{CNN} & \makecell[l]{Camera}
         & \makecell[l]{ $ \textcolor{green}{\raisebox{0.6ex}{\scalebox{0.7}{$\sqrt{}$}}}$ Camera only \\ $ \textcolor{red}{\scalebox{0.85}[1]{$\times$}}$ FoV of Camera \\  $ \textcolor{red}{\scalebox{0.85}[1]{$\times$}}$ No Spatial infomation} 
         & \makecell[l]{  \cite{drews2017aggressive, zheng2019vision}: AutoRally}\\

    \hline
         \textbf{GALNet} & \makecell[l]{Odometry}
         & \makecell[l]{$ \textcolor{green}{\raisebox{0.6ex}{\scalebox{0.7}{$\sqrt{}$}}} $ Fast \\ $ \textcolor{red}{\scalebox{0.85}[1]{$\times$}}$ Dynamic-dependent} 
         & \makecell[l]{\cite{mendoza2020galnet}: Autominy}\\
         
    \hline
         \textbf{LoFTR}  & \makecell[l]{Camera}
         & \makecell[l]{$ \textcolor{green}{\raisebox{0.6ex}{\scalebox{0.7}{$\sqrt{}$}}} $ Camera only \\ $ \textcolor{red}{\scalebox{0.85}[1]{$\times$}}$ High computational demand} 
         & \makecell[l]{ \cite{garcia2023deep}: Jetbot}\\
    \hline 
    \end{tabular}}
\end{table}

\subsubsection{Lane Recognition}
\label{subsec:lane detect}

Lane recognition or lane detection is another important task for the perception module. It is the process of identifying and recognizing the lane markings on a road using algorithms. It typically involves detecting lines or curves that represent the boundaries of lanes on the road. For small-scale cars, we categorize the lane detection methods into traditional and ML-based methods.

\paragraph{Traditional methods}

Traditional lane detection methods rely on computer vision techniques to process sensor data, usually captured by cameras, to identify and track lane markings. Feature extraction techniques are then used to compute metrics such as lateral and orientation deviation, which indicate the position of the car relative to the lane. In \cite{paull2017duckietown}, a multi-step image processing pipeline is employed for lane detection in Duckietown. This pipeline uses techniques such as k-means clustering, the Canny filter \cite{canny1986computational}, HSV colorspace thresholding, and the probabilistic Hough transform \cite{ballard1981generalizing} to extract lane line segments. These features are further processed by a nonlinear non-parametric histogram filter to estimate the lateral displacement and angular offset relative to the right lane center. Similarly, \cite{suto2021real} applies an HSV color-based approach with Gaussian kernel filtering and Sobel edge detection \cite{kanopoulos1988design}, followed by a probabilistic Hough transform for lane detection on the Donkeycar platform. While these traditional methods are computationally efficient and effective for structured environments, their reliance on explicit feature extraction limits their generalization to complex or unstructured scenarios, making them less robust in real-world applications.

\paragraph{ML-based methods}

ML-based approaches on the other hand improve lane detection by leveraging the predictive capabilities of advanced algorithms. For example, \cite{chuang2018deep} employs a Convolutional Neural Network (CNN) to predict lateral displacement and angular offset. Additionally, \cite{zhou2021deep} introduces a modular system trained on a mix of simulation and real-world datasets, mapping sensor inputs into a shared latent space. Recent advancements in Vision Transformers (ViT) have also been applied to lane detection. In \cite{saavedra2022monocular}, a pre-trained ViT is fine-tuned on limited driving datasets to perform environmental segmentation.

\subsubsection{Obstacle Recognition}
\label{subsec:object detect}

Obstacle detection and avoidance refers to identifying and navigating around obstacles or potential hazards in its path to ensure safe and uninterrupted driving behavior. It utilizes various sensors, including cameras, LiDAR, ultrasonic sensors, and other technologies, constantly scan the surroundings to detect and classify obstacles. An effective obstacle detection algorithm necessitates a range of essential abilities. In this paper, obstacle detection algorithms are divided into different groups according to the usage of different sensors, we discuss the most widely used two: camera and LiDAR sensors.

\paragraph{Vision-based approaches}

The vision-based approach relies on images captured by cameras as the primary input for detection. This approach employs computer vision (CV) or ML algorithms, prominently leveraging CNN, for effective detection tasks. Within the realm of cameras used, the vision-based approach can be categorized into two subclasses: monocular and stereo. Monocular image-based methods rely on a single image for processing, while stereo methods utilize images captured by two synchronized cameras. Details about the camera sensors are discussed in Section \ref{subsec:camera}.\par

Within monocular image-based methods, as outlined in \cite{badrloo2022image}, the obstacle detection task for small-scale cars is typically categorized into three primary domains: \emph{appearance-based}, \emph{motion-based}, and \emph{depth-based} methods. With \emph{appearance-based} methods, obstacles are typically identified as the foreground within images. The primary challenge lies in distinguishing relevant foreground or background elements based on established criteria, such as color discrepancies \cite{mashaly2016efficient} or texture features \cite{ulrich2000appearance}. In the work by \cite{nubert2018developing}, the RGB images captured by the monocular camera of a Duckiebot are initially transformed into the HSV colorspace. Subsequently, the color filter from OpenCV is employed to detect obstacles based on color differences. Its reliance on predefined color criteria significantly limits adaptability to diverse scenarios, such as environments with varying lighting or differently colored obstacles. This highlights the need for more robust methods that generalize well across varied settings, such as learning-based techniques. \emph{Motion-based} obstacle detection refers to identifying obstacles or objects in an environment by analyzing motion vectors in the image, it involves comparing successive frames of images to determine alterations in position, velocity, or other motion-related characteristics. Objects that move or exhibit changes in their motion characteristics are then identified as potential obstacles. Optical flow analysis \cite{barron1994performance}, background subtraction, differential methods. can be employed for motion-based obstacle detection \cite{jia2015real, tsai2018vision}. These methods excel in detecting moving obstacles but struggle with static hazards. Moreover, computational demands for real-time implementation can be prohibitive for small-scale platforms, calling for optimization techniques or lightweight motion estimation algorithms. \emph{Depth-based} methods utilize depth information extracted from images to discern the distances and spatial arrangement of objects within the environment, aiding in accurate object detection \cite{kumar2018monocular, hane20173d}. Employing conventional image processing techniques like this may fall short in meeting real-time application expectations, primarily due to their inability to swiftly adapt to dynamic conditions. Therefore, recent research endeavors have pivoted towards enhancing obstacle detection speed by employing CNN and particularly emphasizing the effectiveness of YOLO \cite{redmon2016you}. YOLO is specifically designed for real-time object detection, and its variations have been employed to enhance success rates \cite{dewi2022deep, jiang2022improved, li2023attention}. CNN offers promising capabilities in overcoming the limitations of traditional methods, showcasing greater adaptability and improved real-time performance in diverse and dynamic environments. However, given the restricted computational power available in small-scale cars, a balance must be achieved between accuracy and processing time. In \cite{kannapiran2020go}, YOLOv3 and Tiny YOLO are selected as detection algorithms for the Go-CHART platform. \cite{chang2023accelerating} integrate GhostConv to the YOLOv4-tiny model to achieve faster detection, while in \cite{nguyen2021analysis}, YOLOv5 is used to detect cones and duckies within the Duckietown environment. These aforementioned ML methods still face challenges in balancing detection accuracy and processing time on limited hardware. A potential solution involves further optimizing CNN architectures for edge deployment. Stereo image-based methods work by using a dual-camera system that captures images from slightly different angles, similar to the human binocular vision. This setup helps in perceiving depth and reconstructing three-dimensional scenes by analyzing the differences between the images from these cameras \cite{huh2008stereo}. However, in the context of small-scale cars, constraints such as cost considerations and limited space availability restrict the usage of stereo cameras to select platforms, notably including RACECAR \cite{karaman2017project} and Autominy \cite{berlin2020autominy}. Exploring cost-effective stereo setups, such as single-camera depth estimation augmented with additional processing, could bridge this gap. Vision-based methods offer adaptability and versatility but face challenges related to environmental variability, computational overhead, and hardware limitations. Future work should explore lightweight CNN architectures, as highlighted in \cite{molchanov2017pruning}, where pruning techniques can significantly benefit computation-limited platforms. Additionally, hybrid depth estimation models and motion-aware algorithms hold promise for improving performance within the constraints of small-scale platforms.

\paragraph{LiDAR-based approaches}

In addition to camera-based approaches, LiDAR is also widely used for obstacle avoidance systems. The applications of 2D LiDAR are prevalent in compact car platforms due to factors like size, price, weight, and overall compactness considerations. However, due to the existing computational limitations of these car platforms, innovative LiDAR data processing methods are often unavailable. Consequently, raw LiDAR point cloud data or minimally processed data is frequently used, involving procedures like imputing empty samples and cleansing noisy data, to serve as input for neural networks executing subsequent control tasks. In \cite{morys2023reactive} and \cite{bulsara2020obstacle}, minimal processing of LiDAR data is employed, such as filtering and finding the nearest obstacle, to guide control policies. While computationally efficient, this approach provides limited contextual information, potentially hindering performance in complex scenarios with overlapping or clustered objects. Introducing advanced point cloud processing methods, such as voxel-based filtering or clustering, could enhance obstacle classification and avoidance capabilities. In \cite{o2020f1tenth}, Follow the Gap (FTG) method \cite{sezer2012novel} is employed for object detection and avoidance of an F1TENTH car. This technique involves computing the maximum gap present within the LiDAR point cloud and subsequently determining the steering control command. This approach is effective for reactive obstacle avoidance, especially in dynamic environments. However, it may struggle with narrow gaps. Combining FTG with predictive path planning or vision fusion could improve navigation in cluttered spaces. For small-scale cars, LiDAR remains a cornerstone of obstacle detection, offering reliable distance measurement and robustness. Enhancing LiDAR data processing with ML techniques, such as point cloud segmentation or recurrent neural networks, can address current limitations while maintaining real-time applicability.\par

Both vision-based and LiDAR-based approaches offer unique strengths but are subject to inherent trade-offs. Vision-based methods excel in rich environmental perception and are more cost-effective, but their susceptibility to environmental variability and computational demands can limit their reliability. LiDAR-based methods, while robust and precise, often lack the contextual depth provided by vision systems and can be hindered by computational simplicity. Integrating the two approaches into a multi-modal framework holds significant promise. For instance, vision can provide rich scene context, while LiDAR ensures precise spatial awareness. Techniques like sensor fusion and joint optimization of perception pipelines could leverage the complementary strengths of these modalities.

\begin{table}
    \centering
    \scriptsize
    \caption{Comparison of methods used for obstacle detection.}
    \label{tab:comparision_ob}
    \resizebox{\linewidth}{!}{%
    \begin{tabular}{|l|c|c|l|}
    \hline
          \textbf{Method} & \textbf{Sensor} & \textbf{Analysis} & \textbf{Reference} \\
    \hline
         \textbf{Appearance-based} & \makecell[l]{Camera}&  \makecell[l]{ $ \textcolor{green}{\raisebox{0.6ex}{\scalebox{0.7}{$\sqrt{}$}}}$  Simple, lightweight \\ $ \textcolor{red}{\scalebox{0.85}[1]{$\times$}}$ Lighting or color variations } 
         & \makecell[l]{ \cite{nubert2018developing}: Duckiebot} \\
         
    \hline
         \textbf{Motion-based} & \makecell[l]{Camera}   &  
         \makecell[l]{ $ \textcolor{green}{\raisebox{0.6ex}{\scalebox{0.7}{$\sqrt{}$}}}$  Effective for moving obstacles \\ $ \textcolor{red}{\scalebox{0.85}[1]{$\times$}}$ High computational demand \\ $ \textcolor{red}{\scalebox{0.85}[1]{$\times$}}$ Static obstacles} 
         & \makecell[l]{ \cite{jia2015real,tsai2018vision}: -} \\
         
    \hline
         \textbf{Depth-based} & \makecell[l]{Camera}
         &  \makecell[l]{ $ \textcolor{green}{\raisebox{0.6ex}{\scalebox{0.7}{$\sqrt{}$}}}$ spatial awareness \\ $ \textcolor{red}{\scalebox{0.85}[1]{$\times$}}$ High computational demand \\$ \textcolor{red}{\scalebox{0.85}[1]{$\times$}}$ Dynamic scenarios} 
         & \makecell[l]{\cite{hane20173d, kumar2018monocular}: -}\\

    \hline
         \textbf{CNN-based} & \makecell[l]{Camera}
         & \makecell[l]{ $ \textcolor{green}{\raisebox{0.6ex}{\scalebox{0.7}{$\sqrt{}$}}}$  High adaptability \\ $ \textcolor{red}{\scalebox{0.85}[1]{$\times$}}$ Accuracy and speed trade-off} 
         & \makecell[l]{\cite{kannapiran2020go}: Go-CHART; \cite{chang2023accelerating}: JetRacer;\\ \cite{nguyen2021analysis}: Duckiebot; \cite{yu2022federated}: Jetbot;\\ \cite{elmquist2022performance, de2019cnn}: -}\\

    \hline
         \textbf{Minimum value} & \makecell[l]{LiDAR}
         & \makecell[l]{ $ \textcolor{green}{\raisebox{0.6ex}{\scalebox{0.7}{$\sqrt{}$}}}$ Efficient, robust for nearby objects \\ $ \textcolor{red}{\scalebox{0.85}[1]{$\times$}}$ Limited contextual information} 
         & \makecell[l]{\cite{bulsara2020obstacle,morys2023reactive}: F1TENTH}\\
         
    \hline
         \textbf{FTG} & \makecell[l]{LiDAR}
         & \makecell[l]{ $ \textcolor{green}{\raisebox{0.6ex}{\scalebox{0.7}{$\sqrt{}$}}}$ Effective for reactive obstacle \\ $ \textcolor{red}{\scalebox{0.85}[1]{$\times$}}$ Ineffective for narrow gaps} 
         & \makecell[l]{\cite{o2020f1tenth}: F1TENTH}\\
    \hline
    \end{tabular}}
\end{table}

\subsubsection{Traffic Sign Recognition}
\label{subsec: Traffic sign}

Traffic sign recognition (TSR) is a specialized object detection task that enables autonomous vehicles to interpret road signage and adjust behavior accordingly. Unlike the generic object detection task, TSR uniquely relies solely on cameras, leveraging their capability to capture visual details. The process typically commences with dataset acquisition, wherein cameras capture traffic signs for later utilization. These captured images undergo a series of preprocessing steps to refine quality, eliminate noise, adjust lighting, and employ data augmentation techniques. Subsequently, ML models \cite{sermanet2011traffic, wu2013traffic, satilmics2019cnn} are trained using these compiled datasets for detection and classification. However, research on TSR is limited for small-scale cars due to the unavailability of standard datasets that may differ across platforms. Hence, researchers must prioritize dataset collection as an initial step to train the networks. Future research should intensify exploration in this direction, particularly focusing on mixing TSR with other driving tasks like navigation, lane keeping, or path following, enabling vehicles to make informed, context-aware decisions in real time.

\subsection{Planning}

Path planning in autonomous driving refers to the process by which a self-driving vehicle determines a safe and efficient route from its current location to a desired destination while navigating through its environment \cite{claussmann2019review}. This task involves creating a trajectory or path that the vehicle can follow, considering various factors such as obstacles, road conditions, and traffic regulations, and aims to ensure safe and reliable navigation while optimizing factors like travel time, and energy efficiency. Path planning algorithms vary in complexity and we specify them into traditional \cite{dijkstra1959note, hart1968formal, lavalle2001randomized, li2023adaptive} and ML-based techniques \cite{xiao2022motion, aradi2020survey}.\par

\paragraph{Traditional methods}

Traditional algorithms remain foundational in path planning due to their deterministic nature and efficiency in well-defined scenarios. For example, \cite{liu2021path} applies Dijkstra's algorithm \cite{dijkstra1959note} in conjunction with SLAM methods to extract global map information. The A* algorithm \cite{hart1968formal}, noted for its optimality, is employed in \cite{mccalip2023reinforcement} to plan optimal paths for racing cars, while \cite{sivaprakasam2021improving} integrates Hybrid A* \cite{dolgov2008practical} with learned cost functions to address dynamic obstacles. However, reliance on precomputed cost functions may limit adaptability in novel environments, suggesting the need for adaptive cost-learning techniques. Genetic algorithms (GAs) offer advantages in unstructured environments, as explored in \cite{samadi2013global} and \cite{tuncer2012dynamic}, where stochastic search capabilities generate global paths. However, high computational demands and convergence issues in real-time scenarios could be mitigated by hybrid approaches, combining GAs with deterministic methods like A*. To extend the planning capabilities beyond the sensor horizon, \cite{katyal2021high} utilizes a generative neural network trained on real-world data to predict occupancy maps beyond sensor limitations. This predictive capability assists in facilitating planning processes. Authors leverage the Rapidly-Exploring Random Tree (RRT) algorithm \cite{lavalle2001randomized} to generate a global path, followed by a local planner that orchestrates trajectories until the end of the predicted map. While effective, traditional methods often struggle in dynamic or uncertain environments, emphasizing the importance of more adaptive frameworks.

\paragraph{ML-based methods}

Path planning, treated as an optimization problem, has also seen the successful application of various ML methods, resulting in notably high-performance outcomes. For instance, \cite{gao2019global} uses Q-Learning, a classical RL algorithm, to generate a global path for robots, which achieves a shorter and smoother path compared with the RRT algorithm, though challenges in scalability and training efficiency remain. Advanced reinforcement learning (RL) methods, such as Deep Q-Networks (DQN) and Policy Gradient algorithms, could enhance performance in future studies. A framework is introduced in \cite{zhang2022tvenet} comprising a mapper, global policy, and local policy for image-based navigation for Jetbot. The mapper undergoes supervised learning using camera images to generate an occupancy grid map. Subsequently, a global policy employing DRL techniques takes both the map and the position as input to determine the long-term goal. Following this, a planner, employing the A* algorithm, computes the short-term goal, which is then forwarded to the local planner for further execution. Even with prior maps and acquired information, managing uncertainty during planning and driving remains a significant challenge. To address these uncertainties, \cite{wang2021rough} introduces a model-based RL algorithm incorporating a probabilistic dynamic model. This method aims to mitigate uncertainty during planning stages and prevent shortsighted decisions. In the related approach, \cite{lee2021bayesian} uses Bayesian Residual Policy Optimization (BRPO) with an ensemble of expert policies. Authors train the policy ensemble with BRPO to diminish overall system uncertainty, enabling safe navigation within partially observable environments. Demonstrated on the MuSHR platform, the proposed framework integrates a global localization system, ensuring destination-reaching capabilities while avoiding collisions with other moving vehicles. \par

For path planning, traditional methods excel in static and structured environments due to their computational efficiency and deterministic outputs. However, their inability to adapt to dynamic or uncertain conditions limits their application in complex real-world scenarios. Conversely, ML-based approaches offer flexibility and adaptability, capable of predicting and reacting to unforeseen changes, but they are computationally demanding and heavily reliant on large datasets. Future research should focus on hybrid frameworks that combine the computational efficiency of traditional methods with the adaptability of ML approaches. This could include transfer learning techniques to reduce data dependency, integration of predictive modeling to address uncertainties, and optimization of computational frameworks for real-time applications.

\begin{table}
    \centering
    \scriptsize
    \caption{Comparison of methods used for path planning.}
    \label{tab:comparision_planning}
    \resizebox{\linewidth}{!}{%
    \begin{tabular}{|l|c|l|}
    \hline
          \textbf{Method} & \textbf{Analysis} & \textbf{Reference} \\
    \hline
         \textbf{Dijkstra’s Algorithm} &  \makecell[l]{ $ \textcolor{green}{\raisebox{0.6ex}{\scalebox{0.7}{$\sqrt{}$}}}$  Deterministic \\ $ \textcolor{green}{\raisebox{0.6ex}{\scalebox{0.7}{$\sqrt{}$}}}$ Efficient in static environments \\ $ \textcolor{red}{\scalebox{0.85}[1]{$\times$}}$ Inefficient for large graph } 
         & \makecell[l]{\cite{liu2021path}: -} \\
         
    \hline
         \textbf{A*} &  
         \makecell[l]{ $ \textcolor{green}{\raisebox{0.6ex}{\scalebox{0.7}{$\sqrt{}$}}}$  Efficient in structured environments \\ $ \textcolor{red}{\scalebox{0.85}[1]{$\times$}}$ Struggles in uncertain environment} 
         & \makecell[l]{ \cite{mccalip2023reinforcement}: Deepracer} \\
         
    \hline
         \textbf{Hybrid A*}
         &  \makecell[l]{ $ \textcolor{green}{\raisebox{0.6ex}{\scalebox{0.7}{$\sqrt{}$}}}$ Dynamic scenarios \\ $ \textcolor{red}{\scalebox{0.85}[1]{$\times$}}$ Limited adaptability} 
         & \makecell[l]{\cite{sivaprakasam2021improving}: QCar}\\

    \hline
         \textbf{Genetic Algorithm}
         & \makecell[l]{ $ \textcolor{green}{\raisebox{0.6ex}{\scalebox{0.7}{$\sqrt{}$}}}$  High adaptability \\ $ \textcolor{red}{\scalebox{0.85}[1]{$\times$}}$ High computational demand} 
         & \makecell[l]{\cite{samadi2013global,tuncer2012dynamic}: - }\\

    \hline
         \textbf{RRT}
         & \makecell[l]{ $ \textcolor{green}{\raisebox{0.6ex}{\scalebox{0.7}{$\sqrt{}$}}}$ Efficient in high-dimensional spaces \\ $ \textcolor{red}{\scalebox{0.85}[1]{$\times$}}$ Produce suboptimal paths} 
         & \makecell[l]{\cite{katyal2021high}: RACECAR}\\
         
    \hline
         \textbf{DRL} 
         & \makecell[l]{ $ \textcolor{green}{\raisebox{0.6ex}{\scalebox{0.7}{$\sqrt{}$}}}$ Handles dynamic environments \\ $ \textcolor{red}{\scalebox{0.85}[1]{$\times$}}$ High computational demand} 
         & \makecell[l]{\cite{zhang2022tvenet}: Jetbot; \cite{wang2021rough}: MuSHR;\\ \cite{candela2022transferring}: Duckiebot; \cite{gao2019global}: -}\\

    \hline
         \textbf{BRPO} 
         & \makecell[l]{ $ \textcolor{green}{\raisebox{0.6ex}{\scalebox{0.7}{$\sqrt{}$}}}$ Robust in partially observable environments \\ $ \textcolor{red}{\scalebox{0.85}[1]{$\times$}}$ High computational demand} 
         & \makecell[l]{\cite{lee2021bayesian}: MuSHR}\\ 

    \hline
         \textbf{IL} 
         & \makecell[l]{ $ \textcolor{green}{\raisebox{0.6ex}{\scalebox{0.7}{$\sqrt{}$}}}$ Efficient in known environments \\ $ \textcolor{red}{\scalebox{0.85}[1]{$\times$}}$ Out-of-distribution error} 
         & \makecell[l]{\cite{fukuoka2021self}: Donkeycar}\\ 
    \hline
    \end{tabular}}
\end{table}

\subsection{Control}

Once the path planning module establishes a desired path, the control module becomes the subsequent step. In this section, we will discuss the control methodologies that leverage the reference trajectory to compute control actions to navigate the car.

\begin{table}
    \centering
    \scriptsize
    \caption{Comparison of methods used for the control module.}
    \label{tab:comparision_control}
    \resizebox{\linewidth}{!}{%
    \begin{tabular}{|c|l|l|l|}
    \hline
          \textbf{Control Category} & \textbf{Method} & \textbf{Analysis} & \textbf{Reference} \\
    \hline
         \multirow{11}{*}{\textbf{Path Following}}  & \makecell[l]{\textbf{Kinematic}\\ \textbf{Controller}}&  \makecell[l]{ $ \textcolor{green}{\raisebox{0.6ex}{\scalebox{0.7}{$\sqrt{}$}}}$  Streamlines tuning efforts \\ $ \textcolor{red}{\scalebox{0.85}[1]{$\times$}}$ Relies on accurate system modeling} 
         & \makecell[l]{\cite{hu2021adaptive}: QCar; \cite{bascetta2016kinematic,panahandeh2019self}: -} \\
         
    \cline{2-4}
         & \makecell[l]{\textbf{PID Controller}}   &  
         \makecell[l]{ $ \textcolor{green}{\raisebox{0.6ex}{\scalebox{0.7}{$\sqrt{}$}}}$ Simple, effective in structured systems \\ $ \textcolor{red}{\scalebox{0.85}[1]{$\times$}}$ Limited adaptability } 
         & \makecell[l]{ \cite{paull2017duckietown}: Duckiebot} \\
         
    \cline{2-4}
          & \makecell[l]{\textbf{MPC}}
         &  \makecell[l]{ $ \textcolor{green}{\raisebox{0.6ex}{\scalebox{0.7}{$\sqrt{}$}}}$ Robust in structured environment \\ $ \textcolor{red}{\scalebox{0.85}[1]{$\times$}}$ Requires maps\\$ \textcolor{red}{\scalebox{0.85}[1]{$\times$}}$ Unsuitable for dynamic environment} 
         & \makecell[l]{\cite{spencer2022expert}: MuSHR; \cite{mehrez2013stabilizing,nascimento2018nonlinear,wang2020robust}: -}\\

    \cline{2-4}
          & \makecell[l]{\textbf{IL}}
         & \makecell[l]{ $ \textcolor{green}{\raisebox{0.6ex}{\scalebox{0.7}{$\sqrt{}$}}}$ Adaptable to varied scenarios \\ $ \textcolor{red}{\scalebox{0.85}[1]{$\times$}}$ Struggles with unseen scenarios} 
         & \makecell[l]{\cite{zhang2023zero}:  ART/ATK}\\

    \cline{2-4}
          & \makecell[l]{\textbf{DRL}}
         & \makecell[l]{ $ \textcolor{green}{\raisebox{0.6ex}{\scalebox{0.7}{$\sqrt{}$}}}$ Adaptive to dynamic environment \\ $ \textcolor{red}{\scalebox{0.85}[1]{$\times$}}$ High computational demand} 
         & \makecell[l]{\cite{alomari2021path}: Autominy; \\ \cite{westenbroek2023enabling}: JetRacer}\\
         
    \hline
         \multirow{8}{*}{\textbf{Lane Keeping}} & \makecell[l]{\textbf{PID Controller}}
         & \makecell[l]{ $ \textcolor{green}{\raisebox{0.6ex}{\scalebox{0.7}{$\sqrt{}$}}}$ Simple, effective in structured systems \\ $ \textcolor{red}{\scalebox{0.85}[1]{$\times$}}$ Limited adaptability} 
         & \makecell[l]{\cite{paull2017duckietown, tani2020integrated,chuang2018deep}: Duckiebot}\\
         
    \cline{2-4}
         & \makecell[l]{\textbf{Rule-based}}
         & \makecell[l]{ $ \textcolor{green}{\raisebox{0.6ex}{\scalebox{0.7}{$\sqrt{}$}}}$ Straightforward, efficient \\ $ \textcolor{red}{\scalebox{0.85}[1]{$\times$}}$ High computational demand} 
         & \makecell[l]{\cite{suto2021real}: Donkeycar}\\
         
    \cline{2-4}
         & \makecell[l]{\textbf{IL}}
         & \makecell[l]{ $ \textcolor{green}{\raisebox{0.6ex}{\scalebox{0.7}{$\sqrt{}$}}}$ Adaptable to varied scenarios \\ $ \textcolor{red}{\scalebox{0.85}[1]{$\times$}}$ Struggles with unseen scenarios} 
         & \makecell[l]{\cite{bharadhwaj2019data, perez2019continuous}: Duckiebot; \\ \cite{phan2023robust}: -}\\
         
    \cline{2-4}
         & \makecell[l]{\textbf{DRL}}
         & \makecell[l]{ $ \textcolor{green}{\raisebox{0.6ex}{\scalebox{0.7}{$\sqrt{}$}}}$ Handles dynamic environments \\ $ \textcolor{red}{\scalebox{0.85}[1]{$\times$}}$ Limited adaptability} 
         & \makecell[l]{\cite{li2023vision,li2024platform,beres2023enhancing}: Duckiebot;\\ \cite{zhou2021deep,viitala2021learning}: Donkeycar}\\

    \hline
         \multirow{1}{*}{\textbf{Car Following}} & \makecell[l]{\textbf{DRL}}
         & \makecell[l]{ $ \textcolor{green}{\raisebox{0.6ex}{\scalebox{0.7}{$\sqrt{}$}}}$ Handles dynamic environments \\ $ \textcolor{red}{\scalebox{0.85}[1]{$\times$}}$ Limited adaptability} 
         & \makecell[l]{\cite{li2023vision}: Duckiebot;\\  \cite{bayuwindra2023design}: Donkeycar;\\ \cite{ke2022cooperative}: QCar}\\

    \hline
         \multirow{3}{*}{\textbf{Overtaking}} & \makecell[l]{\textbf{MPC}}
         & \makecell[l]{ $ \textcolor{green}{\raisebox{0.6ex}{\scalebox{0.7}{$\sqrt{}$}}}$ Robust in structured environment \\ $ \textcolor{red}{\scalebox{0.85}[1]{$\times$}}$ Requires maps\\$ \textcolor{red}{\scalebox{0.85}[1]{$\times$}}$ Unsuitable for dynamic environment} 
         & \makecell[l]{\cite{hu2023active}: MuSHR}\\

    \cline{2-4}
         & \makecell[l]{\textbf{DRL}}
         & \makecell[l]{ $ \textcolor{green}{\raisebox{0.6ex}{\scalebox{0.7}{$\sqrt{}$}}}$ Handles dynamic environments \\ $ \textcolor{red}{\scalebox{0.85}[1]{$\times$}}$ Limited adaptability} 
         & \makecell[l]{\cite{li2024platform}: Duckiebot}\\

    \hline
         \multirow{16}{*}{\textbf{Racing}} & \makecell[l]{\textbf{MPC}}
         & \makecell[l]{ $ \textcolor{green}{\raisebox{0.6ex}{\scalebox{0.7}{$\sqrt{}$}}}$ Robust in structured environment \\ $ \textcolor{red}{\scalebox{0.85}[1]{$\times$}}$ Requires maps\\$ \textcolor{red}{\scalebox{0.85}[1]{$\times$}}$ Unsuitable for dynamic environment} 
         & \makecell[l]{\cite{goldfain2019autorally,drews2019vision,wagener2019online}: Autorally;\\  \cite{liniger2015optimization, liniger2017real,liniger2017racing}: ORAC; \\  \cite{jain2020bayesrace}: F1TENTH}\\
         
    \cline{2-4}
         & \makecell[l]{\textbf{MPPI}}
         & \makecell[l]{ $ \textcolor{green}{\raisebox{0.6ex}{\scalebox{0.7}{$\sqrt{}$}}}$ Real-time adaptation \\
         $ \textcolor{red}{\scalebox{0.85}[1]{$\times$}}$ High computational demand} 
         & \makecell[l]{\cite{williams2016aggressive}: Autorally}\\

    \cline{2-4}
         & \makecell[l]{\textbf{Tube-MPC}}
         & \makecell[l]{ $ \textcolor{green}{\raisebox{0.6ex}{\scalebox{0.7}{$\sqrt{}$}}}$ Safe and stable under dynamic conditions \\
         $ \textcolor{red}{\scalebox{0.85}[1]{$\times$}}$ High computational demand} 
         & \makecell[l]{\cite{williams2018robust}: Autorally}\\
         
    \cline{2-4}
         & \makecell[l]{\textbf{LMPC}}
         & \makecell[l]{ $ \textcolor{green}{\raisebox{0.6ex}{\scalebox{0.7}{$\sqrt{}$}}}$ Adaptability to uncertain dynamics \\
         $ \textcolor{green}{\raisebox{0.6ex}{\scalebox{0.7}{$\sqrt{}$}}}$ Improvements with limited information \\ $ \textcolor{red}{\scalebox{0.85}[1]{$\times$}}$ Large datasets \\
         $ \textcolor{red}{\scalebox{0.85}[1]{$\times$}}$ High computational demand} 
         & \makecell[l]{\cite{rosolia2019learning, xue2023learning}: BARC}\\
         
    \cline{2-4}
         & \makecell[l]{\textbf{LPV-MPC}}
         & \makecell[l]{ $ \textcolor{green}{\raisebox{0.6ex}{\scalebox{0.7}{$\sqrt{}$}}}$ Real-time implementation \\
         $ \textcolor{red}{\scalebox{0.85}[1]{$\times$}}$ Limited in highly nonlinear dynamics} 
         & \makecell[l]{\cite{alcala2020autonomous}: BARC}\\

    \cline{2-4}
         & \makecell[l]{\textbf{PAPC}}
         & \makecell[l]{ $ \textcolor{green}{\raisebox{0.6ex}{\scalebox{0.7}{$\sqrt{}$}}}$ Improves safety \\
         $ \textcolor{red}{\scalebox{0.85}[1]{$\times$}}$ High computational demand} 
         & \makecell[l]{\cite{lee2020perceptual}: Autorally}\\

    \cline{2-4}
         & \makecell[l]{\textbf{Pure Pursuit}}
         & \makecell[l]{ $ \textcolor{green}{\raisebox{0.6ex}{\scalebox{0.7}{$\sqrt{}$}}}$ Lightweight, computationally efficient \\
         $ \textcolor{red}{\scalebox{0.85}[1]{$\times$}}$ Less robustness in high-speed} 
         & \makecell[l]{\cite{o2020tunercar}: TUNERCAR}\\

    \hline
         \multirow{3}{*}{\textbf{Drifting}} & \makecell[l]{\textbf{LQR}}
         & \makecell[l]{ $ \textcolor{green}{\raisebox{0.6ex}{\scalebox{0.7}{$\sqrt{}$}}}$ Effective in structured environment \\ $ \textcolor{red}{\scalebox{0.85}[1]{$\times$}}$ Accurate dynamic models required} 
         & \makecell[l]{\cite{gonzales2016autonomous, zhang2017autonomous}: BARC}\\

    \cline{2-4}
         & \makecell[l]{\textbf{NMPC}}
         & \makecell[l]{ $ \textcolor{green}{\raisebox{0.6ex}{\scalebox{0.7}{$\sqrt{}$}}}$ Improves control in dynamic environments \\
         $ \textcolor{red}{\scalebox{0.85}[1]{$\times$}}$ Computational cost} 
         & \makecell[l]{\cite{jelavic2017autonomous}: BARC; \\ \cite{bellegarda2021dynamic}: RACECAR}\\

    \hline
         \multirow{11}{*}{\textbf{Parking}}  
         & \makecell[l]{\textbf{Rule-based}}
         & \makecell[l]{ $ \textcolor{green}{\raisebox{0.6ex}{\scalebox{0.7}{$\sqrt{}$}}}$ Robust for well-defined cases \\
         $ \textcolor{red}{\scalebox{0.85}[1]{$\times$}}$ Limited flexibility} 
         & \makecell[l]{\cite{joung2007study, yi2017smooth, li2021optimization}: -}\\
         
    \cline{2-4}
        &\makecell[l]{\textbf{Fuzzy}}
         & \makecell[l]{ $ \textcolor{green}{\raisebox{0.6ex}{\scalebox{0.7}{$\sqrt{}$}}}$ Simple, interpretable \\ $ \textcolor{red}{\scalebox{0.85}[1]{$\times$}}$ Adaptability in dynamic environments} 
         & \makecell[l]{\cite{chang2002design, li2006autonomous, scicluna2012fpga,li2003autonomous, amarasinghe2007vision, aye2020image, ballinas2018automatic}: -}\\
         
    \cline{2-4}
         & \makecell[l]{\textbf{Neuro-fuzzy}}
         & \makecell[l]{ $ \textcolor{green}{\raisebox{0.6ex}{\scalebox{0.7}{$\sqrt{}$}}}$ Adaptable to diverse scenarios \\
         $ \textcolor{red}{\scalebox{0.85}[1]{$\times$}}$ Training data required} 
         & \makecell[l]{\cite{demirli2009autonomous, wang2010design}: -}\\

    \cline{2-4}
         & \makecell[l]{\textbf{GRBF}}
         & \makecell[l]{ $ \textcolor{green}{\raisebox{0.6ex}{\scalebox{0.7}{$\sqrt{}$}}}$ Dynamic parking environments \\
         $ \textcolor{red}{\scalebox{0.85}[1]{$\times$}}$ Complex implementation} 
         & \makecell[l]{\cite{notomista2017machine}: -}\\

    \cline{2-4}
         & \makecell[l]{\textbf{IL}}
         & \makecell[l]{ $ \textcolor{green}{\raisebox{0.6ex}{\scalebox{0.7}{$\sqrt{}$}}}$ Improves parking precision \\
         $ \textcolor{red}{\scalebox{0.85}[1]{$\times$}}$ High computational demand} 
         & \makecell[l]{\cite{rathour2018vision}: -}\\

    \cline{2-4}
         & \makecell[l]{\textbf{DRL}}
         & \makecell[l]{ $ \textcolor{green}{\raisebox{0.6ex}{\scalebox{0.7}{$\sqrt{}$}}}$ Highly adaptable to dynamic environments \\
         $ \textcolor{red}{\scalebox{0.85}[1]{$\times$}}$ High computational demand} 
         & \makecell[l]{\cite{bejar2019reverse, ozelouglu2022deep}: -}\\

    \hline
         \multirow{5}{*}{\textbf{Cooperative driving}}  
         & \makecell[l]{\textbf{Centralized} \\ \textbf{Control}}
         & \makecell[l]{ $ \textcolor{green}{\raisebox{0.6ex}{\scalebox{0.7}{$\sqrt{}$}}}$ Globally optimized coordination \\
         $ \textcolor{green}{\raisebox{0.6ex}{\scalebox{0.7}{$\sqrt{}$}}}$
         Effective for requiring high precision\\
         $ \textcolor{red}{\scalebox{0.85}[1]{$\times$}}$ High computational demand\\
         $ \textcolor{red}{\scalebox{0.85}[1]{$\times$}}$ Limited scalability} 
         & \makecell[l]{\cite{pohlmann2022ros2}: CoRoLa; \cite{mitchell2020multi}: DeepRacer; \\ \cite{jang2019simulation}: UDSSC;  \cite{swamy2020scaled}: Jetbot; \\ \cite{li2022design}: QCar}\\
         
    \cline{2-4}
        &\makecell[l]{\textbf{Decentralized} \\ \textbf{Control}}
         & \makecell[l]{ $ \textcolor{green}{\raisebox{0.6ex}{\scalebox{0.7}{$\sqrt{}$}}}$ Robustness and scalability \\ $ \textcolor{red}{\scalebox{0.85}[1]{$\times$}}$ Effective communication protocols required \\
         $ \textcolor{red}{\scalebox{0.85}[1]{$\times$}}$ Potential for suboptimal solutions} 
         & \makecell[l]{\cite{hyldmar2019fleet}: CamMini; \cite{sunil2023feature}: QCar; \\ \cite{talia2023pushr}: MuSHR; \cite{scheffe2022increasing}: $\mu$Car; \\\cite{chalaki2020experimental, beaver2020demonstration, chalaki2022research, stager2018scaled}: UDSSC; \\ \cite{garcia2021urban}: Jetbot}\\
     
    \hline
    \end{tabular}}
\end{table}

\subsubsection{Path Following}
\label{subsec:path following}

Path following is the most simple control task for small-scale cars, it necessitates a continual adjustment of the movements based on real-time sensor feedback and environmental alterations, ensuring the vehicle maintains the desired trajectory accurately. Solutions for path following encompass a wide spectrum, including classical control methods like the PID controller, kinematic controller \cite{bascetta2016kinematic, panahandeh2019self, hu2021adaptive}, MPC \cite{mehrez2013stabilizing, nascimento2018nonlinear, wang2020robust, spencer2022expert}, and extending to ML methods \cite{zhang2023zero, alomari2021path, zhang2022tvenet, westenbroek2023enabling}. \par

In \cite{hu2021adaptive}, an adaptive trajectory tracking control scheme is introduced with adjustable gains to facilitate adherence to predefined paths. The designed adaptive control gains aimed to streamline tuning efforts, augment the convergence rate of tracking errors, and elevate trajectory tracing performance. However, a key disadvantage is its reliance on accurate system modeling and gain adjustment, which may struggle under significant environmental disturbances or non-linear dynamics. 
The strategy introduced by \cite{spencer2022expert} involves expert interventions with pre-existing LiDAR-derived maps to further improve performance. Nevertheless, the dependency on pre-existing maps and the requirement for expert input make it less suitable for dynamic or unstructured environments. For ML-based methods, in the work presented by \cite{zhang2023zero}, error states relative to reference trajectories are determined using vehicle state data obtained from sensor fusion involving IMU and GPS with an EKF. A Neural Network (NN) is trained using IL methods with a dataset collected from human-controlled and MPC-controlled driving scenarios. Trained NN maps error states to control commands. While this method significantly reduces the need for complex hand-crafted control laws, it inherits biases from the training data, especially when collected from suboptimal human drivers. To mitigate the need for extensive datasets, \cite{alomari2021path} applies Deep Deterministic Policy Gradient (DDPG) algorithm to guide an Autominy car along the desired path, aiming to minimize cross-track errors. Similarly, in the study by \cite{zhang2022tvenet}, a global policy is determined using DRL, while a local planner processes received images with another DRL agent to derive the final action for the path following. To address the data inefficiency challenges with RL, \cite{westenbroek2023enabling} then introduces a policy gradient-based policy optimization framework leveraging a first-principles model for path following task with a JetRacer. This framework facilitated the learning of precise control policies with limited real-world data. While promising, the reliance on first-principles modeling may limit its adaptability to systems with uncertain or highly non-linear dynamics. Furthermore, the approach assumes access to accurate and reliable simulation environments, which might not always reflect real-world variability. \par

\subsubsection{Lane Keeping}
\label{subsec: Lane Keeping}

When small-scale cars drive on tracks without obstacles, the optimal planned trajectory typically aligns with the center of the right lane. Together with the lane detection methods discussed in Section \ref{subsec:lane detect}, the primary objective is to guide the vehicle within the designated lane, adhering to predefined tolerances for deviations from the lane center. This task requires the car to dynamically adjust its path in real time, based on its position relative to the lane. Consequently, the task heavily depends on visual input, typically utilizing RGB cameras to monitor the lane position of the vehicle continuously. This process, commonly referred to as lane keeping for small-scale vehicles, has been a central focus of extensive research across various platforms \cite{paull2017duckietown, balaji2020deepracer, donkeycar}. Research in lane keeping can be broadly categorized into two approaches: traditional methods and ML-based methods.\par

\paragraph{Traditional methods}

Traditional control methods regulate the trajectory using outputs from the perception module, ensuring it remains within the designated lane. Algorithms such as PID controllers are commonly used to adjust the steering angle based on deviations from the lane center. For instance, in \cite{paull2017duckietown}, the PID controller utilizes lateral displacement and angular offset estimates from the perception pipeline to minimize deviations in real time. Similarly, \cite{suto2021real} integrates a rule-based approach to compute steering angles directly from detected lane lines. These methods are straightforward and interpretable, making them suitable for simple lane keeping tasks. However, their reliance on precise input metrics and static tuning parameters limits their effectiveness in dynamic environments or when system dynamics are nonlinear.

\paragraph{ML-based methods}

ML-based control methods enhance lane keeping by either augmenting traditional control systems or replacing them with more adaptive algorithms. For example, DRL has been successfully used to replace traditional controllers. \cite{li2023vision, li2024platform} employ DRL algorithms to process outputs from the perception module of \cite{paull2017duckietown}, achieving superior performance in multitask driving scenarios beyond lane keeping. Additionally, these systems improve Sim2Real transfer capabilities, as demonstrated by \cite{zhou2021deep}, where a modular system integrates a list of DRL algorithms, utilizing these latent features from the perception module as inputs to control the Donkeycar. While ML-based control methods provide adaptability and can learn optimal policies from data, they are often sensitive to hyperparameter tuning and may require extensive training.

\subsubsection{Car Following}

In addition to navigating tracks without other vehicles, a key driving scenario for autonomous vehicles involves operating alongside other vehicles on the road. One crucial capability in such scenarios is car following, which refers to the ability to maintain a safe and appropriate distance from the vehicle ahead. This requires tracking the movements of the leading vehicle and dynamically adjusting speed and position to ensure safe, comfortable, and efficient driving. The literature on car following maneuvers with small-scale vehicles remains relatively limited, though several notable studies have emerged. A DDPG algorithm with an extended look-ahead approach is proposed in \cite{bayuwindra2023design} for longitudinal and lateral control within vehicle platooning scenarios. For perception, LiDAR technology gauges inter-vehicle distance, while an IMU provides acceleration data. Additionally, a V2V system employing Wi-Fi communication transmits leader information to the follower vehicle. Similarly, in \cite{ke2022cooperative}, a Cooperative Adaptive Cruise Control (CACC) system is implemented utilizing Deep Q-learning (DQN) \cite{mnih2013playing}. This system enables the follower vehicle to dynamically adapt and maintain appropriate inter-vehicular distances. Notably, both these studies rely on V2V communication to achieve effective car following behavior, which increases system complexity and costs. In contrast, \cite{li2023vision} introduces a new perspective by realizing multitasking driving, encompassing car following and lane keeping, solely relying on visual sensors without the need for communication with other vehicles. In this approach, a pattern of circles is affixed behind the leading Duckiebot, aiding the ego Duckiebot in sensing distance and speed to mimic human driver behavior. A modular system is deployed, where the perception module extracts compact affordance information, and a DRL algorithm controls the Duckiebot based on this information. Despite its simplicity, this system yields promising results.

\subsubsection{Overtaking}

Overtaking, like car following, is another critical driving behavior that involves interaction with other vehicles on the track. It refers to the maneuver where an autonomous vehicle changes lanes or positions itself to safely pass another vehicle traveling in the same direction on the road. The purpose of overtaking is to move ahead of the slower vehicle safely and efficiently while maintaining proper traffic flow. For small-scale cars, overtaking is particularly challenging due to the need for precise localization to avoid collisions throughout the maneuver. Consequently, limited research has been conducted on this task. A first contribution comes from \cite{li2024platform}, where authors employ a modular system equipped solely with onboard sensors to execute diverse driving tasks, encompassing lane keeping and overtaking for Duckiebots. The framework leverages traditional machine vision technologies to acquire compact affordance information, as discussed in \cite{paull2017duckietown}, and utilizes LiDAR sensors for distance estimation relative to other vehicles. Afterward, the Long Short-Term Memory Soft Actor-Critic (LSTM-SAC) algorithm assumes the role of the controller. Notably, LSTM aids the agent in recognizing distinct phases within the overtaking process. Demonstrated results affirm the efficacy of this proposed framework, managing both lane keeping and overtaking tasks, and showcasing superior performance when benchmarked against baselines. Another work regarding the overtaking task is \cite{hu2023active}, where a dual control approach with MPC towards active uncertainty reduction is proposed. This approach automatically balances the exploration-exploitation trade-off, enabling the MuSHR car to actively minimize uncertainty concerning the hidden states of other agents without compromising expected planning performance. The difference compared with \cite{li2024platform} from the hardware side is that the vehicle being overtaken will yield, and the usage of a known grid map of the track. While these methods showcase promising results, they highlight the current research limitations in overtaking tasks for small-scale cars. Future work should prioritize the development of more generalizable frameworks capable of handling diverse overtaking scenarios without relying on assumptions such as yielding vehicles or predefined maps.

\subsubsection{Racing}
\label{subsec: Racing}

Autonomous racing, distinct from traditional autonomous driving, emphasizes high-speed navigation, rapid reaction, and dynamic trajectory planning. Unlike standard driving tasks, racing pushes vehicles to attain high velocities, extensively testing the dynamic limits of these automated systems. When racing at high speeds, vehicles must swiftly detect other vehicles or obstacles, demanding rapid reaction times. Additionally, they must accurately localize their position concerning the track and strategize dynamic trajectories to optimize performance \cite{betz2022autonomous}. While extensive research has been dedicated to full-size racing competitions like Roborace, our primary focus centers on small-scale car racing events such as AutoRally \cite{goldfain2019autorally}, F1TENTH \cite{o2020f1tenth}, Donkeycar, ORCA \cite{liniger2015optimization} and others.\par

For racing tasks, the control module has a high demand for timely reactions, often surpassing the importance of the perception module. The most commonly used control module in racing is MPC, which is a control strategy that utilizes a dynamic model of the system to predict its future behavior and make control decisions based on optimization criteria. In \cite{liniger2015optimization}, a path planner and a Nonlinear MPC (NMPC) are utilized to guide the racing process within the ORAC platform. Later, in \cite{williams2016aggressive}, a sampling-based MPC algorithm called the Model Predictive Path Integral Control algorithm (MPPI) is introduced for Autorally. This algorithm presents a novel derivation of path integral control, offering an explicit formula for controls across the entire time horizon. A notable attribute of MPPI is its capability to generate entirely new behaviors dynamically, enabling the controller to drive the vehicle to its operational limits. Then, a few improved versions are proposed, such as the robust sampling-based MPC framework based on a combination of model predictive path integral control and nonlinear Tube-MPC \cite{williams2018robust} and best response model predictive control based on a combination of the game-theoretic notion of iterated best response, and an information-theoretic model predictive control algorithm \cite{williams2018best}. Despite its effectiveness in dynamic trajectory planning and adaptability to high-speed scenarios, MPC has notable drawbacks. It depends heavily on accurate dynamic models, which, if imperfect, can lead to disparities between predicted and actual behavior. Additionally, the high computational demand poses challenges for real-time implementation. 

To address this challenge, Learning Model Predictive Control (LMPC) integrates learning algorithms with MPC, aiming to refine control strategies in the face of uncertain or changing dynamics. Compared to traditional MPC, LMPC continually enhances its performance, especially in scenarios where complete knowledge of system dynamics is unavailable or subject to variation. The learning mechanism algorithms such as RL, Gaussian processes, neural networks, or other adaptive learning methods in LMPC continuously update the predictive model based on collected data and feedback from the system, to achieve better performance, adaptability to changing system dynamics, and robustness in scenarios where precise models might be unavailable or incomplete. \cite{wagener2019online} establishes the connection between MPC and online learning, and proposes a new algorithm based on dynamic mirror descent (DMD) and MPC (DMD-MPC), which provides a fresh perspective on previous heuristics used in MPC. In \cite{rosolia2019learning}, historical data is used to construct secure sets and approximate the value function, facilitating LMPC to learn and improve from past experiences within BARC platform.  Later, \cite{jain2020bayesrace} applies Gaussian processes to correct the model mismatch, then uses MPC for tracking pre-computed racing lines using this corrected model for an F1TENTH car. For more responsive control, \cite{alcala2020autonomous} uses Linear Parameter Varying (LPV) theory to model the dynamics of the vehicle and combine it with MPC (LPV-MPC),  which can be computed online with reduced computational cost. To swiftly identify unsafe conditions, \cite{lee2020perceptual} propose a Perceptual Attention-based Predictive Control (PAPC) algorithm, where MPC is first used to learn how to place attention on relevant areas of the visual input and ROI, and output control actions as well as estimates of epistemic and aleatoric uncertainty in the attention-aware visual input of an Autorally car. \cite{xue2023learning} introduces a local, linear, data-driven learning method for error dynamics within the LMPC framework. This approach exhibits increased robustness against parameter variations and limited data compared to prior LMPC implementations. While LMPC offers significant advantages, it also has limitations. These include high computational costs due to continuous model updates, dependency on sufficient high-quality training data, and the need for additional hardware for onboard real-time learning and processing. Pure pursuit, another control algorithm used in racing tasks, is simpler and more computationally efficient \cite{o2020tunercar}. It is particularly suitable for scenarios with limited computational resources. However, it lacks the adaptability needed for dynamic environments and may struggle in high-speed racing where precise trajectory prediction is essential.\par

One of the significant challenges in autonomous racing is localization. Current methods, such as motion capture systems and pre-mapped tracks, rely heavily on external infrastructure \cite{liniger2018path, chisari2021learning}. This dependence creates scalability issues and imposes hardware and environmental constraints. To overcome these limitations, future research should focus on developing map-free or vision-based localization systems. Lightweight neural networks could also be leveraged for real-time learning and decision-making, reducing the reliance on extensive infrastructure. \par

Although autonomous racing does not commonly occur in daily driving scenarios, it is still an emerging field in intelligent vehicles and transportation systems. The intriguing aspect lies in the operation of autonomous small-scale cars pushing the boundaries of vehicle capabilities \cite{betz2022autonomous}. Operating at high speeds with minimal reaction time within dynamic environments, autonomous racing with small-scale cars emerges as a compelling area within the autonomous driving domain.

\subsubsection{Other tasks}

In addition to the control tasks discussed earlier, we also explore tasks that are less commonly studied for small-scale cars, broadening the scope of research in this domain.

\paragraph{Drifting}

Autonomous drifting, often associated with high-performance racing, represents a complex and sophisticated challenge in autonomous vehicle control. It requires precise coordination of speed, steering, throttle, and braking, alongside an in-depth understanding of vehicle dynamics and kinematics. Effective state estimation methods are equally critical to ensure stability and accuracy during drifting maneuvers. Traditional approaches heavily rely on dynamic models, such as the dynamic bicycle model used in \cite{gonzales2016autonomous}, which employs EKF to fuse sensor data and integrates a Linear-Quadratic Regulator (LQR) controller with equilibrium drifting points for control. While this method is effective within its design constraints, its performance is tightly coupled to the accuracy of the dynamic model and the predefined operating conditions. Similarly, \cite{zhang2017autonomous} expands upon this by introducing a six-state bicycle model, yet it struggles with generalizing to more complex and dynamic scenarios. This highlights a fundamental challenge: precise vehicle dynamics models are often impractical or unattainable in real-world environments. To address these limitations, \cite{jelavic2017autonomous} proposes an approach to tackle the drifting park problem with BARC cars by segmenting it into distinct phases: the normal driving regime and the sliding regime. During the normal driving phase, a nonlinear MPC operates with a predefined kinematic model. As the vehicle transitions into the sliding phase, reliance on this model diminishes. To navigate this shift, a feedforward-feedback controller takes charge, orchestrating safer maneuvers adeptly under sliding conditions. While effective, this segmentation introduces additional computational complexity and coordination challenges. Moreover, tire model singularities at low speeds exacerbate control difficulties, especially in the sliding regime. In response to these challenges, \cite{bellegarda2021dynamic} presents a novel solution by integrating a fused kinematic-dynamic bicycle model with a nonlinear MPC framework tailored for a RACECAR platform.  This unified approach harmonizes the planning and execution of dynamic vehicle maneuvers within a unified framework, optimizing the entire process cohesively. Despite these advancements, conventional dynamic models are constrained by their reliance on simplified assumptions, limiting their applicability. Future research should prioritize integrating data-driven methods, such as neural network-based controllers or DRL controllers, to improve adaptability and performance in uncertain environments. Additionally, the development of robust state estimation techniques that can operate effectively with noisy sensor inputs and less precise dynamic models is crucial for advancing autonomous drifting capabilities.

\paragraph{Parking}
\label{subsec:parking}

Besides driving behaviors, parking is also a crucial research area in AD for both small-scale and normal-scale cars. Research on the car parking problem generally stems from the broader motion planning problem and is typically defined as finding a collision-free path that connects the initial configuration to the final one. Traditional parking methods generally follow a three-phase approach: mapping the parking space, planning a collision-free path, and executing the maneuver. For small-scale cars, fuzzy logic algorithms dominate traditional methods. These systems rely on distance sensors, such as ultrasonic, infrared \cite{chang2002design, li2006autonomous, scicluna2012fpga}, or LiDAR \cite{li2003autonomous, amarasinghe2007vision, aye2020image, ballinas2018automatic}, to detect parking spaces and use predefined fuzzy rules to determine steering angles. While these methods are simple and effective in structured environments, they lack adaptability and struggle with complex, dynamic scenarios. Closed-loop controllers and rule-based strategies, while robust for specific cases, also suffer from limited flexibility \cite{joung2007study, yi2017smooth, li2021optimization}. \par

ML-based methods offer adaptability and learning capabilities, making them better suited for dynamic parking environments. Neuro-fuzzy systems combine neural networks with fuzzy logic \cite{jang1997neuro} to dynamically update rules and membership functions based on training data \cite{demirli2009autonomous, wang2010design}, improving performance in diverse scenarios. Other ML approaches segment the parking process for better control, such as the use of a General Radial Basis Function (GRBF) classifier and Random Forest kernel to identify behavior transitions, as seen in \cite{notomista2017machine}. Later, \cite{rathour2018vision} proposed a two-stage learning framework to predict steering angles and gear status for parking using front and back-mounted monocular cameras. In the first stage, an encoder-decoder architecture estimates the initial steering angle trajectory. This trajectory, along with the heading angle and absolute position, is then fed into an LSTM network to estimate the optimal steering angle and gear status for parking. DRL approaches such as DDPG and other end-to-end learning frameworks have also shown promise in learning complex parking maneuvers, incorporating environmental feedback for precise control \cite{bejar2019reverse, ozelouglu2022deep}. While ML methods excel in adaptability, they often require large datasets, extensive training, and computational resources. \par

Despite the importance of autonomous parking for small-scale cars, research in this area has been lacking over the past few decades. However, with the development of current ML methods, there is a growing need to employ more advanced techniques in this research topic.

\paragraph{Cooperative Driving}
\label{subsec: Cooperative driving}

Compared to the previously discussed scenarios, which typically involve a single controlled vehicle, cooperative driving entails a collaborative approach where multiple autonomous vehicles communicate, with V2V communications.  In this setting, vehicles interact to achieve shared goals, navigate complex environments, and enhance traffic efficiency and safety. This involves exchanging data such as positions, speeds, and intended trajectories to enable collective decision-making, which is generally categorized into centralized and decentralized control strategies.\par

In centralized control, a central entity coordinates the actions of all vehicles by processing shared data to make globally optimized decisions. This approach excels in tasks like collaborative SLAM and multi-vehicle path planning, where precise coordination is crucial. For instance, \cite{sunil2023feature} utilizes feature-based map registration for collaborative SLAM among three vehicles via V2V communication, enhancing mapping speed and accuracy. The centralized strategy excels in achieving global optimization and consistency but is limited by scalability challenges and potential single points of failure. The dependency on robust communication infrastructure also makes it vulnerable to latency and data synchronization issues. In contrast, decentralized control distributes decision-making among vehicles, which act based on local sensor data and peer communication. An example is the decentralized MPC method in \cite{talia2023pushr}, where MuSHR cars plan collision-free trajectories through negotiation, enabling precise and independent control. Safety is further addressed by \cite{scheffe2022increasing}, which proposes a priority-based distributed MPC (P-DMPC) algorithm to minimize collision risks. The decentralized control is advantageous for scalability and fault tolerance but can face challenges in achieving global optimization due to limited information sharing and potential inconsistencies among individual vehicle actions.\par 

Specific applications of cooperative driving highlight its versatility. In roundabout navigation, vehicles communicate and negotiate entry, adjusting speeds and trajectories to yield to those already in the roundabout. RL-based coordination methods like those used in \cite{jang2019simulation} and decentralized control frameworks from \cite{chalaki2020experimental} demonstrate smooth merges and reduced stop-and-go traffic in scaled testbeds. Similarly, traffic jam mitigation emphasizes synchronized vehicle actions to maintain optimal flow, as seen in frameworks like \cite{hyldmar2019fleet}, which adapts the IDM and MOBIL models for cooperative behavior, and studies like \cite{beaver2020demonstration} and \cite{chalaki2022research}, which reduce travel times through decentralized optimal control. Emerging areas extend beyond autonomous cars. For example, \cite{pohlmann2022ros2} achieves adaptive cruise control for multi-vehicle platooning, while works such as \cite{li2022design} explore cooperation between UGVs and UAVs, pushing the boundaries of collaborative traffic management.\par

For future works, we propose to include hybrid control systems that combine centralized coordination with decentralized adaptability, robust V2V communication systems that minimize latency and security risks, and cross-platform collaboration between diverse vehicle types to enhance traffic management and operational flexibility. These advancements promise to unlock the full potential of cooperative driving in real-world applications.

\subsection{End-to-end Driving}

After discussing the modular system, we then delve into the end-to-end systems for small-scale cars. Here, we primarily focus on two main streams: lane keeping and racing. \par

For the lane keeping task, the end-to-end system has received more extensive research compared to modular systems. In this paradigm, the agent processes raw sensor input and outputs control commands directly, bypassing the modular perception, planning, and control layers. This framework primarily encompasses two methods: IL and DRL. Among IL methods, BC is foundational due to its simplicity, as it maps observed states directly to corresponding actions without modeling the underlying decision-making process. For example, in \cite{verma2021implementation, podbucki2022aspects}, CNNs are trained with image data to  predict directly the steering angle for control under various evaluation metrics within the DeepRacer platform. BC offers the advantages of straightforward implementation and computational efficiency, making it effective in scenarios where expert behavior is well-defined and consistent. However, its drawbacks include vulnerability to compounding errors, where deviations from expert behavior escalate in unseen states, and a heavy dependence on the quality and diversity of training data, which often hinders its ability to generalize to out-of-distribution scenarios. To address these issues, enhancements to BC have been proposed. For instance, \cite{stocco2022mind} introduces uncertainty modeling during training to improve Sim2Real transfer on the Donkeycar platform. Similarly, \cite{liu2020data} augments datasets using image style transfer \cite{gatys2016image} to enhance generalization. Comparative studies, such as \cite{lHorincz2022imitation}, evaluate BC, GAIL, and DAgger in the Duckietown environment, highlighting the potential of Domain Adaptation techniques for better Sim2Real transfer. IDRL, in contrast, learns control policies directly from raw sensor data. Studies such as \cite{balaji2020deepracer, dreveck2021easy, revell2022sim2real} employ raw images as input to DRL controllers, while preprocessing methods like dimensionality reduction are used in \cite{almasi2020robust, wiggers2021learning, kalapos2020sim} to improve convergence speed. Addressing control smoothness, \cite{cao2023image} introduces Conditioning for Action Policy Smoothness (CAPS), which reduces jerky control behaviors. Sim2Real transfer capabilities are improved by including delays and sampling rate as additional observations during training, as demonstrated by \cite{sandha2021sim2real}, enhancing the robustness of DRL policies on the DeepRacer platform. Beyond image inputs, raw LiDAR data serves as input in studies such as \cite{ivanov2020case}, where DRL algorithms navigate F1TENTH cars in structured environments. Although DRL eliminates the need for expert demonstrations and is highly adaptable to complex and dynamic environments, it often suffers from high computational cost and data inefficiency. Additionally, ensuring smooth and stable control remains challenging, particularly in real-world scenarios, and achieving reliable Sim2Real transfer is difficult due to discrepancies (Sim2Real gap) between training and deployment environments. To further enhance generalization and convergence, autoencoders are frequently integrated into end-to-end systems \cite{rumelhart1986learning}. By compressing input data into lower-dimensional latent representations, autoencoders provide an efficient representation for downstream tasks such as control or policy learning. For example, \cite{bharadhwaj2019data, perez2019continuous} preprocess image data with autoencoders before training Duckiebots using BC with expert trajectories. Variational Autoencoders (VAEs) \cite{kingma2013auto} have also been explored in DRL, as demonstrated in \cite{viitala2021learning, beres2023enhancing}, where compressed latent features facilitate faster convergence and improved learning efficiency compared to raw image inputs. Despite their advantages in computational efficiency and feature extraction, autoencoders risk losing critical information during compression, particularly in dynamic or noisy environments, and introduce additional computational overhead during training and integration. \par

For racing with end-to-end systems, IL and DRL \cite{hamilton2022zero} have been widely explored. The authors in \cite{pan2017agile} introduce an end-to-end IL system, where a learner network needs to imitate an expert. The expert fuses GPS and IMU for state estimation and uses an MPC as controller, the learner uses a DNN as control policy to map raw, high-dimensional observations to continuous steering and throttle commands. Building on this, \cite{cai2021vision} presents a deep imitative RL framework for end-to-end racing with camera input, where IL is used to initialize the policy, and model-based RL is used for further refinement by interacting with an uncertainty-aware world model. This hybrid approach achieves improved adaptability and performance by leveraging the strengths of both IL and RL. To improve the stability of the racing system, \cite{lee2019ensemble} designs a residual control system, in which multiple Bayesian Neural Networks (BNNs) are trained with two camera inputs and GPS measurements to control an Autorally car in an end-to-end fashion. The integration of LiDAR technology has further expanded the capabilities of end-to-end systems. In \cite{brunnbauer2022latent}, a model-based RL approach effectively utilizes raw LiDAR input to navigate racetracks with an F1TENTH car. Following this, \cite{bosello2022train} employs LiDAR data as input for a DQN to control the F1TENTH car, offering a comparative analysis of neural network architectures for LiDAR data processing. This study also evaluates two Sim2Real approaches, demonstrating the importance of transferability for real-world deployment. Despite the demonstrated potential of DRL for racing tasks, the well-known issue of low sample efficiency continues to hinder its broader application.  For this issue, \cite{zhang2022residual} propose an efficient residual policy learning method with the raw observation of LiDAR and IMU, in which first a controller based on the modified artificial potential field (MAPF) is used to generate policies, then DRL algorithms are used to generate a residual policy as a supplement to obtain the optimal policy with increased efficiency. Similarly, \cite{evans2023high} presents a trajectory-aided learning (TAL) method that trains DRL with raw LiDAR input by incorporating the optimal trajectory into the learning formulation. \cite{trumpp2023residual} also presents a residual vehicle controller that learns to amend a traditional controller with a similar idea. To tackle the safety issues that usually occur in RL training, \cite{evans2023safe} uses a Viability Theory-based supervisor to recursively feasible vehicle safety during the training.  Furthermore, to improve the robustness of RL, \cite{chu2020sim} first train a teacher model that overfits the training track, moving along a near-optimal path, then use this model to teach a student PPO model the correct actions along with randomization.

Looking ahead, hybrid approaches that combine IL and DRL for end-to-end systems show significant promise. Pretraining policies with IL to capture expert knowledge, followed by DRL fine-tuning for adaptability, can improve both sample efficiency and performance. Furthermore, advanced Sim2Real techniques, such as generative models and adversarial training, are critical for bridging the gap between simulated and real-world environments. Exploring multi-modal input fusion, and integrating diverse sensors like images, LiDAR, and radar, could enhance robustness and adaptability. Additionally, incorporating interpretable ML techniques would make the decision-making process of end-to-end systems more transparent, a crucial step for safety-critical applications. Finally, developing real-time adaptation algorithms that allow models to learn and adjust online without requiring retraining could ensure more reliable performance in dynamic, real-world settings. By addressing these challenges and pursuing these research directions, end-to-end systems can become more robust, efficient, and scalable, enabling broader adoption in real-world autonomous driving scenarios.

\subsection{Proposed Framework}

After reviewing the benchmarked tasks and techniques, it is evident that ML and DL methods form the backbone of AD research for small-scale cars. These methods play an important role across key modules such as perception, planning, and control. In perception, techniques like CNNs and autoencoders enhance object detection, lane detection, and semantic segmentation. For control, IL and RL enable models to make efficient and safe decisions in diverse and dynamic conditions. Connected driving leverages the Internet of Things (IoT) to create an interactive ecosystem, enabling real-time data exchange via V2V and V2I communication. This supports coordinated behaviors like platooning and optimized traffic flow management. \par

We would like to propose a baseline framework that integrates these methods into a cohesive structure. By extracting an effective framework from the literature, we aim to provide valuable insights for researchers. In a modular pipeline system, the process begins with the perception layer. Sensor fusion combines data from multiple sensors such as cameras and LiDAR using DNN, IMU, GPS, and encoders using KF alongside SLAM techniques. This integration ensures robust environmental perception and creates a comprehensive understanding of the environment. The fused feature or raw sensor input is then used for object detection, employing ML or shallow learning models to identify and classify objects in the vehicle's vicinity. Compared with deep ML methods, shallow models have simple structures, usually consisting of one or a few layers of processing units, but are effective for resource-constrained environments, in our case, small-scale cars with lower-end computing units. Support vector machines or K-Nearest Neighbors can be used for perceptions. Subsequently, path planning and behavior planning are performed, either separately or within a single module. Classical methods like A*, Dijkstra's, and RRT can be utilized for path planning, while behavior planning can leverage finite state machines, DRL, IL, MPC, and decision trees to handle tasks such as lane keeping, lane changing, overtaking, and obstacle avoidance. A safety and redundancy module ensures system reliability, incorporating PID or rule-based controllers for safe stopping during failures and redundant systems for robustness. For end-to-end systems, raw sensor data is processed directly using IL and RL or their hybrid approaches for driving tasks, with safety modules remaining integral. This cohesive framework aims to guide researchers in advancing small-scale autonomous driving systems.

\section{Future trends}
\label{sec:Future}

In the previous sections, we have gone through the currently available small-scale car platforms, their hardware configurations, and the autonomous driving tasks achieved by these platforms. In pursuit of augmenting the capacity and applicability of these small-scale car platforms across diverse age groups, facilitating educational use for students, and supporting sophisticated research purposes, we list the following opportunities for further exploration.\par

\subsection{Enhancing Accessibility and Usability}

It is imperative to contemplate transforming the small-scale platform into a more widely accessible resource for educational purposes and academic research. This necessitates the establishment of a lower entry barrier, reduced pricing, and the implementation of a comprehensive learning pipeline. For educational usage, the first consideration should be given to the ease of learning and maintenance for teachers. A more user-friendly starting point will likely foster increased enthusiasm among educators, encouraging them to incorporate the platform into their daily teaching activities. This, in turn, will provide students with a challenging learning experience, ultimately preparing them for the intricacies of fully autonomous driving in the near future. For research applications, an in-depth exploration of the accessibility of the small-scale car platform is essential. This entails tailoring entry levels to accommodate various research goals, catering to individuals with zero experience in robotics to seasoned professionals. The platform should not only serve as a testbed for autonomous systems but also ignite enthusiasm for exploring diverse robotic configurations. While some platforms currently offer support for different entry levels, a more comprehensive development is warranted to meet the diverse needs of the research community.\par

\subsection{Improving Versatility and Advanced Technology Adoption}
Regarding the more academic research consumption, the versatility of the platforms should be more considered. Specifically, concerning individual small-scale cars, attention should be directed towards enhancing their functional dimensions. In line with advancements in semiconductor technology, where computation and sensor units are becoming more compact yet powerful, platforms should incorporate more sophisticated sensors, thereby augmenting the overall capabilities of the system. As discussed in Section \ref{sec:Task}, most of the current techniques applied in small-scale car platforms are old methods developed for full-scale cars, with simplified apply conditions. With the rapid advancement of AD research, increasingly sophisticated techniques in sensing, perception, and control are being developed for full-scale vehicles. These novel methods typically require advanced sensor suites, including 3D LiDAR and cameras, which demand significant computational resources and power supply—requirements that most small-scale car platforms cannot meet. Additionally, deploying these techniques in real-world systems necessitates fast communication between components. As a result, more powerful platforms are becoming increasingly attractive. \par

\subsection{Bridging Gaps in Smart City Configurations}
We propose a thorough examination of smart city configurations. While smart cities hold significant potential in advancing research on autonomous driving for small-scale cars, there exists a gap in terms of accessibility, reproducibility, and standardization of best practices. There is an urgent need for a common framework across the research community, encompassing both hardware and software aspects. By addressing these two aspects, small-scale car platforms can more closely resemble real-world AD systems. It unlocks the potential for deploying widely used open-source AD systems, such as Apollo and Autoware, on small-scale car platforms. These AD systems necessitate a comprehensive ecosystem encompassing sensing, perception, localization, planning, and control. Furthermore, existing smart city setups often overlook key elements such as weather conditions and pedestrian interactions, which are important in real-world driving scenarios. Addressing these factors in future research is essential for developing comprehensive and realistic autonomous driving solutions within smart city environments.\par

\subsection{Advancing V2V and V2I Integration}

Achieving fully autonomous driving necessitates the integration of critical technologies such as V2V, V2I, and V2X. While these technologies have been extensively researched in the literature, there remains a noticeable gap in discussions of small-scale car platforms. Consequently, the community needs to pivot towards exploring and advancing V2V and V2I communications as the next steps in the pursuit of comprehensive autonomous driving solutions.

\section{Conclusion}
\label{sec:Conclusion}

In this survey, we offer an overview of the current state-of-the-art developments in small-scale autonomous cars. Through an in-depth exploration of past and ongoing research, we identify critical challenges and highlight the promising trajectory for advancing small-scale autonomous driving technology. We begin by enumerating the presently predominant small-scale car platforms widely employed in academic and educational domains, detailing the configurations and specifications of each. Similar to their full-scale counterparts, the deployment of hyper-realistic simulation environments is imperative for training, validating, and testing autonomous systems before real-world implementation. To this end, we show the commonly employed universal simulators and platform-specific simulators. Furthermore, we provide a detailed summary and classify the literature into distinct categories: end-to-end systems versus modular systems and traditional methods versus ML-based methods. This classification facilitates a nuanced understanding of the diverse approaches adopted in the field. We introduce methods used for perception, path planning, control, and end-to-end driving. To provide a holistic guide for researchers and practitioners, we also outline the commonly utilized components and tools across various well-known platforms. This information serves as a valuable resource, enabling readers to leverage our survey as a guide for constructing their own platforms or making informed decisions when considering commercial options within the community.\par

We additionally present future trends concerning small-scale car platforms, focusing on different primary aspects. Firstly, enhancing accessibility across a broad spectrum of enthusiasts: from elementary students and colleagues to researchers, demands the implementation of a comprehensive learning pipeline with diverse entry levels for the platform. Next, to complete the whole ecosystem of the platform, a powerful car body, varying weather conditions, and communications issues should be addressed in a smart city setup. These trends are anticipated to shape the trajectory of the field, contributing significantly to advancements in real-world autonomous driving research. \par   

While we have aimed to achieve maximum comprehensiveness, the expansive nature of this topic makes it challenging to encompass all noteworthy works.  Nonetheless, by illustrating the current state of small-scale cars, we hope to offer a distinctive perspective to the community, which would generate more discussions and ideas leading to a brighter future of autonomous driving with small-scale cars.
\addtolength{\textheight}{-1.1cm}
\section*{Acknowledgments}
This work was funded by ScaDS.AI (Center for Scalable Data Analytics and Artificial Intelligence) Dresden/Leipzig.


\bibliographystyle{IEEEtran}

\bibliography{ref}

\end{document}